\documentclass{article}

 \usepackage[preprint]{neurips_2020}

\usepackage{docmute}
\usepackage[utf8]{inputenc} 
\usepackage[T1]{fontenc}    
\usepackage{url}            
\usepackage{booktabs}       
\usepackage{amsfonts}       
\usepackage{nicefrac}       
\usepackage{microtype}      

\usepackage{graphicx}
\usepackage{paralist, amsmath, amsthm, amssymb, graphicx, hyperref}
\usepackage{amssymb}
\usepackage{setspace}
\usepackage{mathtools}
\usepackage{nccmath}
\usepackage{xcolor}

\usepackage{nicefrac}
\usepackage{stmaryrd}
\usepackage{bm}
\usepackage{cancel}
\usepackage{multirow}

\usepackage{xfrac}
\usepackage{mathrsfs}
\usepackage{amscd}
\usepackage{dsfont}
\usepackage{tikz}
\usepackage{epstopdf}
\usepackage{float}
\usepackage{enumerate}
\usepackage{bm}
\usepackage{chngcntr} 
\usepackage{makecell}
\usepackage{placeins}
\usepackage{algorithm}
\usepackage{algorithmic}
\usepackage{multicol}

\usepackage{xr}
\usepackage{hyperref}

\newcommand{\la}{\lambda}
\newcommand{\sa}{\sigma}

\renewcommand{\l}{\left}
\renewcommand{\r}{\right}

\newcommand{\norm}[1]{\left\lVert#1\right\rVert}
\newcommand{\normbig}[1]{\big\lVert#1\big\rVert}
\newcommand{\abs}[1]{\l|#1\r|}

\newcommand{\red}{\color{red}}

\DeclareMathOperator{\prox}{prox}

\theoremstyle{plain}

\newtheorem{prop}{Proposition}

\title{An Efficient Semi-smooth Newton Augmented \\ Lagrangian Method for Elastic Net}

\author{%
  Tobia Boschi \thanks{\texttt{tub37@psu.edu}}\\
  Department of Statistics\\
  Penn State University \\
  University Park, USA\\
  \And
  Matthew Reimherr \\
  Department of Statistics \\
  Penn State University \\
  University Park, USA \\
  \And 
  Francesca Chiaromonte \thanks{and EMbeDS, Sant'Anna School of Advanced Studies, Pisa, Italy}  \\
  Department of Statistics \\
  Penn State University \\
  University Park, USA \\
}

\begin{document}

\maketitle


\begin{abstract}
Feature selection is an important and active research area in statistics and machine learning. The Elastic Net is often used to perform selection when the features present non-negligible collinearity or practitioners wish to incorporate 
additional known structure.
In this article, we propose a new Semi-smooth Newton Augmented Lagrangian Method to efficiently solve the Elastic Net in ultra-high dimensional settings. Our new algorithm exploits both the sparsity induced by the Elastic Net penalty and the sparsity due to the second order information of the augmented Lagrangian. This greatly reduces the computational cost of the problem. Using simulations on both synthetic and real datasets, we demonstrate that our approach outperforms its best competitors by at least an order of magnitude in terms of CPU time. We also apply our approach to a Genome Wide Association Study on childhood obesity. 
\end{abstract}

%
%

\section{Introduction}
\label{sec:intro}

The advent of big data, with applications involving massive numbers of predictors, has made feature selection a central research area in statistics and machine learning. Many regularization approaches have been developed to solve this problem, such as Lasso \citep{tibshirani1996regression}, SCAD \citep{fan2001variable}, adaptive Lasso \citep{zou2006adaptive}, constrained Lasso \citep{gaines2018constrained}, etc.
While effective in many settings, Lasso has strong limitations in scenarios characterized by very high collinearities among features. To tackle this issue, \citet{zou2005regularization} introduced the Elastic Net, which penalizes both the $l_1$ and the squared $l_2$ norm of the coefficients; the former induces sparsity, while the latter regularizes coefficient estimates mitigating variance inflation
due to collinearity.
The Elastic Net minimization is formulated as 
\begin{equation}
\small
\label{eq:objective_function}
    \min_x \frac{1}{2} \norm{Ax - b}_2^2 + \la_1 \norm{x}_1 + \frac{\la_2}{2} \norm{x}_2^2,
\end{equation}
where $n$ and $m$ are the number of features and observations, respectively, $b \in \mathbb{R}^m$ is the response vector, $A \in \mathbb{R}^{m\times n}$ is the standardized design matrix, and $x \in \mathbb{R}^n$ is the coefficient vector. 
Many algorithms exists to efficiently solve \eqref{eq:objective_function}, such as accelerated proximal gradient \citep{li2015proxgrad}, FISTA \citep{beck2009fast},  distributed ADMM \citep{boyd2011distributed} and coordinate descent \citep{tseng2009coordinate, friedman2010regularization} -- see \citet{boyd2004convex} for a more exhaustive list. 
We develop a new \emph{Semi-smooth Newton Augmented Lagrangian} method 
for the Elastic Net (\texttt{SsNAL-EN}) in ultra-high-dimensional settings -- where the number of features is much larger than the number of observations.
Other versions of SsNAL have been recently introduced by \citet{li2018} to solve regular Lasso, and then extended to constrained Lasso by \citet{deng2019}. 

\texttt{SsNAL-EN} exploits the sparsity induced by the second order information of the dual augmented Lagrangian to dramatically reduce computational costs.
Moreover, the structure of the Elastic Net penalty guarantees a super-linear convergence of both the augmented Lagrangian algorithm and its inner sub-problem, just as in the Lasso case.
Therefore, very few iterations are needed to solve both instances with high accuracy. 

We implemented 
an efficient version of our method in \texttt{python} 
and benchmarked it 
against 
two different versions of the coordinate descent algorithm -- one implemented in the \texttt{python package sklearn} and one implemented in the \texttt{R package glmnet}. \texttt{glmnet} is written in \texttt{fortran} and is considered the 
gold standard for fitting sparse generalized linear models. 
We benchmarked our method also against three advanced solvers which implement screening rules to improve computational performance: the \texttt{R package biglasso} (written in \texttt{C}$++$), 
and the \texttt{python packages Gap Safe Rules} \citep{ndiaye2017gap} and \texttt{celer} \citep{massias2018celer}.
Simulation results demonstrate the 
 comparative efficiency of our new approach.

We also applied \texttt{SsNAL-EN} to the Intervention Nurses Start Infants Growing on Healthy Trajectories (INSIGHT) study
\citep{paul2014intervention, craig2019polygenic}, which examines risk factors for childhood obesity. 
We investigate the association between single nucleotide polymorphisms (SNPs) and two scalar outcomes -- a conditional weight gain score (CWG) and Body Mass Index (BMI) -- detecting SNPs that may affect obesity risk in children. 

The remainder of the article is organized as follows. In Section \ref{sec:prel}, we introduce some preliminaries on Fenchel conjugate functions and proximal operators. 
In Section \ref{sec:methodology}, 
we describe our new 
method.
In Section \ref{sec:sim_and_case}, we present 
simulations on both synthetic data and real datasets, and we apply our method to the INSIGHT study. In Section \ref{sec:conclusions}, we provide final remarks and discuss future developments. 
The \texttt{python} code for \texttt{SsNAL-EN} and 
for our simulations is available at \href{https://github.com/tobiaboschi/ssnal_elastic}{github.com$/$tobiaboschi$/$ssnal$\_$elastic}. Data from the INSIGHT study is sensitive and privacy protected, and hence not provided.

%
%

\section{Preliminaries}
\label{sec:prel}

\begin{figure}[]
	\begin{center}
	\centerline{\includegraphics[width=0.8\linewidth]{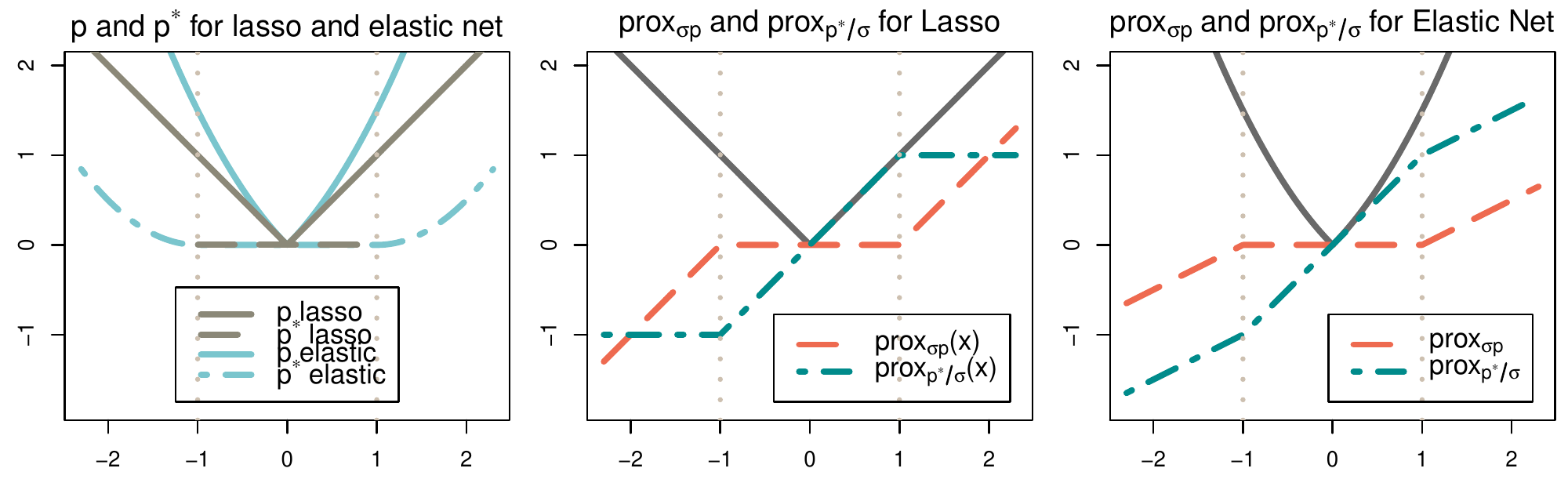}}
	\vspace{-0.4cm}
	\caption{The left panel shows 
	penalty functions (solid lines) and their conjugate functions (dotted lines) for the Lasso (brown) and the Elastic Net (blue). The 
	central panel shows $\prox_{\sa p}$ (red line) and $\prox_{p^*/\sa}$ (green line) for the Lasso.
	The right panel shows $\prox_{\sa p}$ (red line) and $\prox_{p^*/\sa}$ (green line) for the Elastic Net. 
	In all panels the vertical dotted lines represent the interval $[-\la_1, \la_1]$, and we consider $\la_1 = \la_2 = \sa = 1$ and $x, z \in \mathbb{R}$. }   	
	\label{fig:prelimaries}
	\end{center}
	\vspace{-0.6cm}
\end{figure}
\texttt{SsNAL-EN} relies on \emph{Fenchel conjugate functions} \citep{fenchel1949conjugate} and \emph{proximal operators} \citep{rockafellar1976augmented, rockafellar1976} -- which are 
well know in the optimization field and used in a wide range of problems and applications. In this section, we briefly introduce these
mathematical objects, providing definitions and basic properties.


\subsection{Fenchel conjugate functions}
\label{subsec:Fenchel}

Fenchel conjugate functions (or Legendre transformations) allow one to easily define the dual problem and the Lagrangian dual function \citep{boyd2004convex}. %
Let $\mathcal{X} \in \mathbb{R}^n$ be a convex set and $f~:~\mathcal{X} \rightarrow \mathbb{R}$. The conjugate function of $f$ is $f^*~:~\mathcal{X}^* \rightarrow \mathbb{R}$ defined as $f^*(z) = \sup_{x \in \mathcal{X}} \big(z^T x - f(x) \big)$, 
where $\mathcal{X}^* = \big\{ z \in \mathbb{R}^n~:~ \sup_{x \in \mathcal{X}} \left(z^T x - f(x) \right) < \infty \big\}$. 
As an example, the conjugate function of the Lasso penalty $p(x) = \la_1 \norm{x}_1$ is
\begin{equation}
\small
    p^*(z) =  \mathds{1}_{\{\norm{z}_\infty \le \la \}} = 
    \begin{cases}
        0 &\norm{z}_\infty \le \la_1 \\
        \infty & o.w.
    \end{cases}. 
\end{equation}
%
Our first result 
gives a closed form 
for the Elastic Net penalty conjugate function, 
which is pivotal to solve the augmented Lagrangian problem. 
\begin{prop}
\label{prop:p_star}
    Let 
    $p(x)= \la_1 \norm{x}_1 + (\la_2/2) \norm{x}_2^2$ be the Elastic Net penalty, then 
    \begin{equation}
    \small
    \label{eq:p_star}
        p^*(z) = 
        \frac{1}{2\la_2} \sum_{i=1}^n 
            \begin{cases}
                (z_i - \la_1)^2, & z_i \ge \la_1 \\
                0, & \abs{z_i} < \la_1 \\
                (z_i + \la_1)^2, & z_i \le -\la_1 \\
            \end{cases}.
    \end{equation}
\end{prop}
A proof of this proposition, which follows the same lines as that of \citet{dunner2016primal}, is provided in Supplement \ref{app:proof_p_star}. 
The left panel of Figure \ref{fig:prelimaries} depicts Lasso and Elastic Net penalties and their conjugate functions when $x, z \in \mathbb{R}$. While for the Lasso  $p^*$  is an indicator function, for the Elastic Net $p^*$ is a continuous differentiable function equal to $0$ in the interval $[-\la_1, \la_1]$.


\subsection{Proximal operators}
\label{subsec:proximal}

Let $f~:~\mathbb{R}^n \rightarrow \mathbb{R} $ be a lower semi-continuous convex function. 
The proximal operator of $f$ at $x$ with parameter $\sigma >0$, denoted as $\prox_{\sigma f}: \mathbb{R}^n \rightarrow \mathbb{R}^n$, is defined as 
\begin{equation}
\label{eq:prox_op}
\small
  \prox_{\sigma f}(x) = \arg \min_t \big( f(t) + (2\sigma)^{-1} \norm{t- x}^2_2 \big)
\end{equation}
\noindent
\citet{parikh2014proximal} interpret 
this as an approximated \textit{gradient step} for $f$. Indeed, when $f$ is differentiable and $\sa$ is sufficiently small, we have $\prox_{\sa f}(x) \approx x - \nabla f(x)$. To implement \texttt{SsNAL-EN}, we need the proximal operator of the penalty function $p$ and of its conjugate $p^*$. Given the first, 
the second is obtained through the \emph{Moreau decomposition}: $ x = \prox_{\sa p}(x) + \sa \prox_{\frac{1}{\sa}p^*}(x/ \sa)$ for $\sa > 0$. 

As an example, the proximal operator of the Lasso penalty $p(x) = \la_1 \norm{x}_1$ is the soft-tresholding operator \citep{parikh2014proximal}. In particular, for each component $i = 1,\ldots n$ of $x$, we have
\begin{equation}
\small
\label{eq:prox_l1}
    \prox_{\sigma p}(x_i) =
        \begin{cases}
            x_i - \sigma\la_1, & x_i \ge \sigma\la_1 \\
            0, & \abs{x_i} < \sigma\la_1 \\
            x_i + \sigma\la_1, & x_i \le - \sigma\la_1 \\
        \end{cases},
    \phantom{i}
    \prox_{p^*/\sa}(x_i/ \sa) = 
        \begin{cases}
            \la_1, & x_i \ge \sigma\la_1 \\
            x_i/ \sa, & \abs{x_i} < \sigma\la_1 \\
            -\la_1, & x_i \le - \sigma\la_1 \\
        \end{cases} 
\end{equation}
To obtain the proximal operator of the Elastic Net penalty $p(x) = \la_1 \norm{x}_1 + (\la_2/2)\norm{x}^2_2$,  
one composes $\prox_{(\sa \la_2/2) \norm{\cdot}_2^2}$ and the soft-thresholding operator obtaining
\begin{equation}
\small
	\label{eq:prox_elastic}
		\prox_{\sigma p}(x_i) = \frac{1}{1 + \sa \la_2} 
        	\begin{cases}
                x_i - \sigma\la_1, & x_i \ge \sigma\la_1 \\
                0, & \abs{x_i} < \sigma\la_1 \\
                x_i + \sigma\la_1, & x_i \le - \sigma\la_1 \\
            \end{cases},
        \phantom{i}
        \prox_{p^*/\sa}(x_i/ \sa) = 
        \begin{cases}
            \frac{(x_i \la_2 + \la_1)}{(1 + \sa \la_2)}, & x_i \ge \sigma\la_1 \\
            x_i/ \sa, & \abs{x_i} < \sigma\la_1 \\
            \frac{(x_i \la_2 - \la_1)}{(1 + \sa \la_2)}, & x_i \le - \sigma\la_1 \\
        \end{cases}
\end{equation}
The 
central and right panels of Figure \ref{fig:prelimaries} depict $\prox_{\sa p}(x)$ and $\prox_{p^*/\sa}(x)$ for the Lasso and Elastic Net penalties, respectively, when $x \in \mathbb{R}$. Just as in the Lasso case, the Elastic Net proximal operator induces sparsity in the interval $[-\la_1, \la_1]$. Outside this interval, the Elastic Net operator still grows linearly, but with a slope smaller than the Lasso one, due to the presence of the scaling factor $\la_2$.

%
%

\section{Methodology}
\label{sec:methodology}

In this section we describe our new method and its 
implementation.
\texttt{SsNAL-EN} focuses on the augmented Lagrangian of the dual formulation of \eqref{eq:objective_function}, exploiting the sparsity of its second order information to greatly reduce computational cost. 

Consider the continuous differentiable function $h(Ax)=(1/2)\norm{Ax-b}_2^2$ and the closed proper function $p(x) = \la_1 \norm{x}_1 + (\la_2/2)\norm{x}^2_2$.
The Elastic Net minimization \eqref{eq:objective_function} can be expressed as
\begin{equation}
\small
\label{eq:primal}
\tag{P}
    \min_x \left(h(Ax) + p(x) \right).
\end{equation}
From \citet{boyd2004convex}, a possible dual formulation of \eqref{eq:primal} is
\begin{equation}
\small
\label{eq:dual}
\tag{D}
    \min_x - \left(h^*(y) + p^*(z)\right) ~ | ~ A^T y + z = 0, 
\end{equation}
where $y \in \mathbb{R}^m$ and $z \in \mathbb{R}^n$ are the dual variables, and $h^*$ and $p^*$ are the conjugate functions of $h$ and $p$. In particular, $h^*(y) = (1/2)\norm{y}_2^2 + b^T y$ \citep{dunner2016primal} and $p^*(z)$ is given in Proposition \ref{prop:p_star}. The Augmented Lagrangian function \citep{fenchel1949conjugate} associated with \eqref{eq:dual} is
\begin{equation}
\small
\label{eq:lagrangian}
    \mathcal{L}_\sa (y, z, x) = h^*(y) + p^*(z) - x^T \big(A^T y + z \big) + (\sa / 2) \normbig{A^T y + z}_2^2.
\end{equation}
where $x \in \mathbb{R}^n$ is the \emph{Lagrange multiplier} and penalizes the dual constraint's violation.
To study the optimality of the primal and dual problems and the convergence of our method, we also introduce the \emph{Karush-Kuhn-Tucker} (KKT) conditions associated with \eqref{eq:dual}, which are:
\begin{equation}
\small
\label{eq:kkt}
        \nabla h^*(y) - Ax = 0,  \quad
        \nabla p^*(z) - x = 0, \quad  
        A^T y + z = 0,
\end{equation}
where $\nabla h^*(y) = y -b$. Finding the closed form of $\nabla p^*(z)$ is not essential for our method.
\citet{boyd2004convex} 
proved that $(\bar y, \bar z, \bar x)$ solves the KKT \eqref{eq:kkt} if and only if $(\bar y, \bar z)$ and $(\bar x)$ are the optimal solutions of \eqref{eq:dual} and \eqref{eq:primal}, respectively. 


\subsection{The augmented Lagrangian problem}
\label{subsec:augmented_lagrangian}

\begin{algorithm}[tb]
\small 
\caption{Semi-smooth Augmented Lagrangian (SnNAL) method}
\label{alg:al}
\begin{multicols}{2}
\begin{algorithmic}
	\STATE {\bf Augmented Lagrangian Method}
	\vspace{0.1cm}
	\STATE Start from the initial values $y^0, z^0, x^0, \sa^0$
	\vspace{0.1cm}
	\WHILE{not convergence}
	\vspace{0.1cm}
	\STATE {\bf(1)} Given $x^i$, find $y^{i+1}$ and $z^{i+1}$ which approximately solve the inner sub-problem
		\begin{equation}
		\small
        \label{eq:inner_subproblem}
         	 (y^{i+1}, z^{i+1}) \approx \arg \min_{y,z} \mathcal{L}_\sa (y, z ~|~ x^i)
        \end{equation}
    \vspace{-0.3cm}
	\STATE {\bf(2)} Update the Lagrangian multiplier $x$ and the parameter $\sa$:
   		\begin{align}
   		\small
        \label{eq:update_x}
                    \begin{split}
                        &x^{i+1} = x^i - \sa_k (A^T y^{i+1} + z^{i+1}) \\
                        &\sa^{i+1} \uparrow \sa^{\infty} \le \infty
                    \end{split}
        \end{align}
    \vspace{-0.2cm}
    \ENDWHILE
    \STATE $\phantom{i}$
    \STATE $\phantom{i}$
    \STATE {\bf Semi-smooth Newton method}
	\vspace{0.1cm}
    \STATE To solve the sub-problem \eqref{eq:inner_subproblem} and find  $(y^{i+1}, z^{i+1})$:
	\vspace{-0.2cm}
	\WHILE{not convergence}
	\vspace{0.1cm}
	\STATE {\bf(1)} Find descent direction $d^j$ solving exactly or by \emph{conjugate} \emph{gradient} the linear system:
		\begin{equation}
		\small
                \label{eq:newton_direction}
                    \partial^2 \psi (y^j) d^j = -\nabla \psi (y^j)
        \end{equation}
   \vspace{-0.4cm}
   \STATE {\bf(2)} \emph{Line search} \citep{li2018}: choose $\mu \in (0, 1/2)$ and  reduce the step size $s^j$ until:
   		\begin{equation}
   		\small
        \label{eq:line_search}
        	\psi (y^j + s^j d^j ) \le \psi ( y^j) + \mu s^j \langle \nabla (y^j), d^j \rangle
   		\end{equation}
   	\vspace{-0.3cm}
   	\STATE {\bf(3)} Update $y$: $y^{j+1} = y^j + s^j d^j$
   	\vspace{0.1cm}
   	\STATE {\bf(4)} Update $z$: $z^{j+1} = \prox_{p^*/\sa} \left( x^i/\sa^i - A^T y^{j+1} \right)$
   	\vspace{-0.2cm}
    \ENDWHILE
\end{algorithmic}
\end{multicols}
\vspace{-0.3cm}
\end{algorithm}
As described in \citep{rockafellar1976augmented}, one can find the optimal solution of \eqref{eq:dual} by solving the \emph{Augmented Lagrangian} (AL) method described in {\bf Algorithm \ref{alg:al}}.
The critical part here is solving the inner sub-problem \eqref{eq:inner_subproblem}. Based on \citet{li2018}, for a given $x$, an approximate solution $\left( \bar y, \bar z \right)$ can be computed simultaneously as 
\begin{equation}
\small
\label{eq:z_bar_y_bar}
    \bar y = \arg \min_y \mathcal{L}_\sa \left(y ~|~ \bar z, x \right), \quad 
    \bar z = \arg \min_z \mathcal{L}_\sa \left(z ~|~ \bar y, x \right). 
\end{equation}
This leads to our second result (a proof again is provided in Supplement \ref{app:proof_psi}). 
\begin{prop}
\label{prop:psi_y_z_bar}
Define $\psi(y) := \mathcal{L}_\sa \left(y ~|~ \bar z, x \right)$. Then, for the Elastic Net 
we have:
    \begin{align}
    \small
    \label{eq:psi_y_z_bar}
    \begin{split}
        \small
        &(1)~\psi(y) = h^*(y) + \frac{1 + \sa \la_2}{2 \sa} \normbig{\prox_{\sa p}\big( x - \sa A^T y\big) }_2^2 - \frac{1}{2\sa} \norm{x}_2^2 \\ 
        &(2)~\bar z =  \prox_{p^*/\sa} \big( x/\sa - A^T \bar y \big) \\ 
    \end{split}
    \end{align}
\end{prop}
\noindent
$\prox_{\sa p}$ and $\prox_{p^*/\sa}$ are given in \eqref{eq:prox_elastic}.
Note that $\psi$ is a continuous differentiable function. As we will see in more detail next, in order to solve \eqref{eq:inner_subproblem} one has to minimize $\psi$ with respect to $y$ or, equivalently, find the solution of $\nabla \psi(y) = 0$.


\subsection{A Semi-smooth Newton method to solve (17)}
\label{subsec:semismooth_newton}

To solve the augmented Lagrangian sub-problem \eqref{eq:inner_subproblem}, we propose the \emph{Semi-smooth Newton} (SsN) method described in {\bf Algorithm \ref{alg:al}}, where $\hat \partial^2 \psi (y)$ denotes the generalized Hessian of $\psi$ at $y$.
SsN updates $y$ and $z$ iteratively; $z$-updates follow the rule in Proposition \ref{prop:psi_y_z_bar} and $y$-updates consist of minimizing $\psi$ through one Newton step. 
The main challenge is the computational cost of solving the linear system \eqref{eq:newton_direction} -- which we substantially reduce exploiting the sparse structure of $\partial^2 \psi (y)$.  
Since $\psi$ is continuous and differentiable, its gradient is
\begin{equation}
\small
\label{eq:grad_phy}
    \nabla \psi (y) = \nabla h^*(y) - A \prox_{\sa p}\big(x - \sa A^T y\big),
\end{equation}
where $\nabla h^*(y) = y + b$. We can define the operator
\begin{equation}
\small
\label{eq:hessian_hat}
    \hat \partial^2 \psi (y) := \nabla^2 h^*(y) + \sa  A \partial  \prox_{\sa p}\big(x - \sa A^T y \big) A^T,
\end{equation}
where $\nabla^2 h^*(y) = I_m$ (the $m \times m$ identity matrix) and $\partial  \prox_{\sa p}$ is the Clarke subdifferential \citep{clarke1990optimization}.
If we choose $Q \in \partial  \prox_{\sa p} (x - \sa A^T y)$, then $V := I_m + \sa A Q A^T \in \hat \partial^2 \psi (y)$. Moreover, from \citet{hiriart1984hessian}, we have $\hat \partial^2 \psi (y) d = \partial^2 \psi (y) d$ for every $d$ in the domain of $y$. It follows that, as long as $Q$ is properly chosen, solving \eqref{eq:newton_direction} is equivalent to solving $V d = -\nabla \psi \left(y\right)$.

We now illustrate the pivotal role of $Q$ in \texttt{SsNAL-EN} and how it can induce sparsity in the linear system \eqref{eq:newton_direction}. 
Let $Q$ be the $n \times n$ diagonal matrix with entries
\begin{equation}
\small
\label{eq:Q_def}
    q_{ii} = \frac{1}{1 + \sa \la_2}
        \begin{cases}
            1 & \left|\left(x - \sigma A^T y \right)_i\right| > \sigma \la_1 \\
            0 & \text{o.w.} 
        \end{cases} .
\end{equation}
It is easy to verify that $Q \in \partial  \prox_{\sa p} (x - \sa A^T y)$. Let $\mathcal{J} = \left\{j~:~ | (x - \sigma A^T y )_i | > \sigma \la_1 \right\}$, and $r = |\mathcal{J}|$ be the cardinality of $\mathcal{J}$. 
Given the structure of $Q$, we have $\sa A Q A^T = \kappa A_\mathcal{J} A_\mathcal{J}^T$, where $\kappa =  \sa / \l(1 + \sa \la_2\r)$ and $A_\mathcal{J} \in \mathbb{R}^{m \times r}$ is the sub-matrix of $A$ restricted to the columns in $\mathcal{J}$. 
The system \eqref{eq:newton_direction} thus becomes:
\begin{equation}
\small
\label{eq:AJ_system}
    \big(I_m + \kappa A_\mathcal{J} A_\mathcal{J}^T \big) d =  -\nabla \psi (y).
\end{equation}
Note $V$ is positive semidefinite because both $I_m$ and $A Q A^T$ are. 
Using the \emph{Cholesky factorization} the total cost of solving the linear system reduces from $\mathcal{O} \l(m^2(m+n)\r)$ to $\mathcal{O} \l(m^2(m+r)\r)$. 
This includes computing $A_\mathcal{J} A_\mathcal{J}^T$ ($\mathcal{O} \l(m^2 r\r)$) and the Cholesky factorization ($\mathcal{O} \l(m^3 \r)$). 
Because of the sparsity induced by the Elastic Net, $r$ is usually much smaller than $n$ -- causing substantial computational gains. Even when $n$ is very large $\l( \sim 10^7 \r)$, one can still solve the linear system efficiently.
Furthermore, if $r<m$, which is often the case when the Elastic Net solution is sparse, one can factorize an $r \times r$ (instead of $m \times m$) matrix using the \emph{Sherman-Morrison-Woodbury} formula: 
%
\begin{equation}
\small
\label{eq:smw_formula}
	    \big(I_m + \kappa A_\mathcal{J} A_\mathcal{J}^T \big)^{-1} = 
	    I_m - A_\mathcal{J}  \big(\kappa^{-1} I_r +A_\mathcal{J}^T A_\mathcal{J} \big)^{-1} A_\mathcal{J}^T. 
\end{equation}
In this case the total cost of solving the linear system is further reduced from $\mathcal{O} \l(m^2(m+r)\r)$ to $\mathcal{O} \l(r^2(m+r)\r)$. 
To achieve this efficiency, we leverage both the sparsity produced by the penalty and the sparsity inherent to the second order operator $\hat \partial^2 \psi (y)$.
Finally, if in the first iterations of the algorithm $m$ and $r$ are both larger than $10^4$, we can further improve computing performance by solving \eqref{eq:newton_direction} approximately with the \emph{conjugate gradient} method. 

A full convergence analysis for both the \emph{Augmented Lagrangian} and the \emph{Semi-smooth Newton} method is provided in Supplement \ref{app:convergence}. Given the super-linear rate, both methods require just a few iterations to converge, as we will see in Section \ref{sec:sim_and_case}. In practice, to determine the convergence of AL and SnN, we check the residuals of the third and first KKT in \eqref{eq:kkt}, respectively, i.e:
\begin{equation}
\small
\label{eq:kkt_res}
    \text{res}(\text{kkt}_3) = (1 + \norm{y}_2 + \norm{z}_2)^{-1}\normbig{A^T y + z}_2, \quad
    \text{res}(\text{kkt}_1) = (1 + \norm{b}_2)^{-1} \norm{y + b - Ax}_2
\end{equation}
\noindent

\subsection{Parameter tuning}
\label{subsec:tuning_par}

To guide the choice of  $\l( \la_1, \la_2 \r)$, we consider three different quantitative criteria: 
\emph{k-fold Cross Validation} ($cv$) \citep{tibshirani1996regression}, \emph{Generalized Cross Validation} ($gcv$) \citep{jansen2015generalized}, and \emph{Extended Bayesian Information Criterion} ($e$-$bic$) \citep{chen2012extended} -- which modifies the standard BIC to also include the number of features $n$. 
While effective in a wide range of scenarios, $cv$ can be very computationally expensive because it requires solving $k$ additional Elastic Net problems for each value of $\l( \la_1, \la_2 \r)$. 
In contrast, $gcv$ and $e$-$bic$ are computed directly from the original solution as:
\begin{equation}
\label{eq:gcv_bic}
\small
	\text{gcv}(\hat x) = (m)^{-1} \text{rss} (\hat x) / \l( 1 - \nu /m \r)^2 \quad 
	\text{e-bic}(\hat x)  = \log \l(\text{rss}(\hat x)/m\r) + (\nu /m) \l( \log m + \log n \r) 
\end{equation}
where $\text{rss}(\hat x)$ is the residual sum of squares associated with the solution $\hat x$, and $\nu$ are the Elastic Net degrees of freedom. 
Indicating with $\mathcal{J}$ the active set of $\hat x$, we have $\nu = \text{tr} ( A_\mathcal{J} ( A_\mathcal{J}^T A_\mathcal{J} + \la_2 I_r) ^{-1} A_\mathcal{J}^T)$ \citep{tibshirani2012degrees}.
Before computing the criteria, we de-bias Elastic Net estimates by fitting standard least squares on the selected features. Albeit naive, this approach is effective when $n \gg m$ \citep{belloni2014debiasedols, zhao2017debiasedols}. 
To ensure efficiency of the parameter tuning component of our code, we implement
some important refinements. 
We start from values of $\la_1$ very close to $\lVert A^Tb \rVert_{\infty}$, i.e.~the smallest value of $\la_1$ which gives a solution with 0 active components. These values are associated to very sparse solutions, which are fast to compute. When we move to the next value of $\la_1$, we use the solution at the previous value for initialization (\emph{warm-start}): the new solution is very close to the previous one and can be computed very quickly -- usually \texttt{SsNAL-EN} converges in just one iteration. 
Finally, we allow the user to fix the maximum number of active features: when this number is reached, no further $\la_1$ or $\la_2$ values are explored.

%
%

\section{Simulation study and INSIGHT data}
\label{sec:sim_and_case}

%
%
%
\begin{figure}[]
	\vspace{0.2cm}
	\begin{center}
	\centerline{\includegraphics[width=0.97\linewidth]{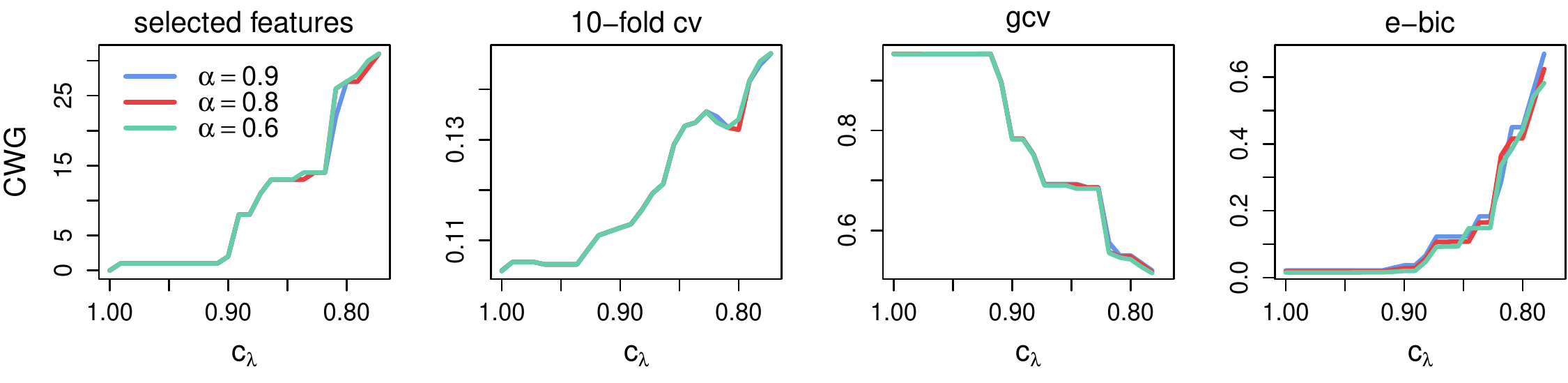}}
	\vspace{-0.4cm}
	\centerline{\includegraphics[width=0.97\linewidth]{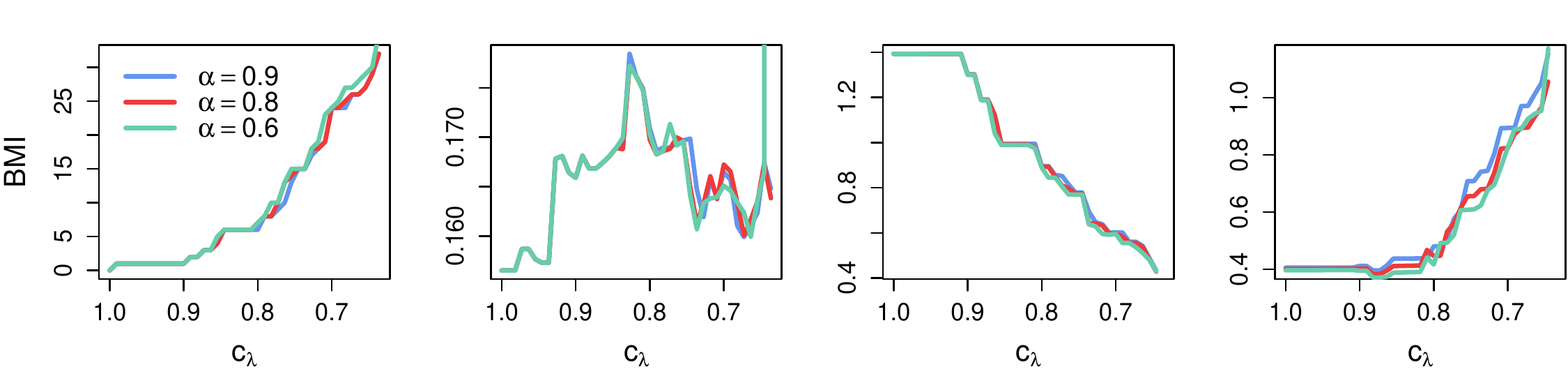}}
	\vspace{-0.4cm}
	\caption{
	Parameter tuning criteria for
	the INSIGHT data as $c_\la$ varies on the horizontal axis. 
	Top and bottom row refer to the CWG and BMI regressions, respectively. For each, from left to right, we have: number of selected features, \emph{10-fold cv}, \emph{gcv}, and \emph{e-bic}.
	We consider three values of $\alpha$: $0.9$ (blue line), $0.8$ (red line), $0.6$ (green line). 
	}    
	\label{fig:cwg_tuning}
	\end{center}
	\vspace{-0.6cm}
\end{figure}
In this section, we demonstrate the gains provided by \texttt{SsNAL-EN} on simulated data and on some commonly used reference data sets. Furthermore, we employ our new method to perform feature selection in a study of genetic variants associated to childhood obesity.


\subsection{Simulation settings and results}
\label{subsec:sim_settings}

\begin{table}[]
\caption{CPU time (in seconds) of \texttt{glmnet}, \texttt{sklearn} and \texttt{SsNAL-EN} for different values of $n$ and different simulation scenarios. For \texttt{SsNAL-EN}, we also report the number of iterations in parentheses.} 
\small
\begin{sc}
\centerline{
\scalebox{0.9}{
\begin{tabular}{rr|rrr|rrr|rrr}
\Xhline{3\arrayrulewidth}
\multicolumn{1}{l}{} & \multicolumn{1}{l|}{} & \multicolumn{3}{c|}{{\bf SIM 1}} & \multicolumn{3}{c|}{{\bf SIM 2}} & \multicolumn{3}{c}{{\bf SIM 3}} \\ 
\Xhline{3\arrayrulewidth}
\multicolumn{1}{r|}{$\bm n$} & $\bm{\hat\rho}$ & \multicolumn{1}{c}{\emph{glmnet}} & \multicolumn{1}{c}{\emph{sklearn}} & \multicolumn{1}{c|}{\emph{ssnal-en}} & \multicolumn{1}{c}{\emph{glmnet}} & \multicolumn{1}{c}{\emph{sklearn}} & \multicolumn{1}{c|}{\emph{ssnal-en}} & \multicolumn{1}{c}{\emph{glmnet}} & \multicolumn{1}{c}{\emph{sklearn}} & \multicolumn{1}{c}{\emph{ssnal-en}} \\ 
\Xhline{3\arrayrulewidth}
\multicolumn{1}{r|}{$1e4$} & 1.4 & 0.084 & 0.116 & $\bm{0.026}(4)$ & 0.074 & 0.129 & $\bm{0.031}(4)$ & 0.067 & 0.071 & $0.010(4)$ \\
\multicolumn{1}{r|}{$1e5$} & 1.1 & 1.174 & 1.113 & $\bm{0.157}(3)$ & 0.834 & 0.940 & $\bm{0.153}(4)$ & 0.734 & 0.896 & $\bm{0.109}(4)$ \\
\multicolumn{1}{r|}{$5e5$} & 1.0 & 3.615 & 4.869 & $\bm{0.607}(3)$ & 3.696 & 4.129 & $\bm{0.841}(4)$ & 3.671 & 6.147 & $\bm{0.517}(4)$ \\
\multicolumn{1}{r|}{$1e6$} & 1.0 & 22.644 & 29.399 & $\bm{1.311}(3)$ & 7.173 & 9.312 & $\bm{1.792}(4)$ & 7.783 & 10.079 & $\bm{1.192}(4)$ \\
\multicolumn{1}{r|}{$2e6$} & 1.0 & 97.031 & 134.247 & $\bm{3.188}(3)$ & 88.216 & 140.378 & $\bm{2.995}(4)$ & 71.763 & 132.738 & $\bm{2.360}(4)$ \\ 
\Xhline{3\arrayrulewidth}
\end{tabular}}}
\end{sc}
\label{tab:times}
\end{table}
\begin{table}[]
\caption{CPU time (in seconds) of \texttt{glmnet}, \texttt{sklearn} and \texttt{SsNAL-EN} for 
the reference data sets. For \texttt{SsNAL-EN}, we also report the number of iterations in parentheses.}
\small
\begin{sc}
\centerline{
\scalebox{0.9}{
\begin{tabular}{cc|rrr|rrr|rrr}
\Xhline{3\arrayrulewidth}
\multicolumn{1}{l}{} & \multicolumn{1}{l|}{} & \multicolumn{3}{c|}{$\bm{housing8}$} & \multicolumn{3}{c|}{$\bm{bodyfat8}$} & \multicolumn{3}{c}{$\bm{triazines4}$} \\
\multicolumn{2}{c|}{$(m;$ $n;$ $\hat \rho)$} & \multicolumn{3}{c|}{$(506; 203489; 103)$} & \multicolumn{3}{c|}{$(252; 319769; 193)$} & \multicolumn{3}{c}{$(186; 557844; 27)$} \\ 
\Xhline{3\arrayrulewidth}
\multicolumn{1}{c|}{$\bm\alpha$} & $\bm r$ & \multicolumn{1}{c}{\textit{glmnet}} & \multicolumn{1}{c}{\textit{sklearn}} & \multicolumn{1}{c|}{\textit{ssnal-en}} & \multicolumn{1}{c}{\textit{glmnet}} & \multicolumn{1}{c}{\textit{sklearn}} & \multicolumn{1}{c|}{\textit{ssnal-en}} & \multicolumn{1}{c}{\textit{glmnet}} & \multicolumn{1}{c}{\textit{sklearn}} & \multicolumn{1}{c}{\textit{ssnal-en}} \\ 
\Xhline{3\arrayrulewidth}
\multicolumn{1}{c|}{\multirow{2}{*}{0.8}} & 20 & 1.715 & 27.836 & $\bm{0.464}(4)$ & 1.423 & 56.848 & $\bm{0.707}(5)$ & 1.743 & 51.043 & $\bm{1.267}(6)$ \\
\multicolumn{1}{c|}{} & 5 & 1.673 & 3.269 & $\bm{0.204}(2)$ & 1.362 & 9.039 & $\bm{0.235}(3)$ & 1.640 & 16.728 & $\bm{0.917}(5)$ \\
\hline
\multicolumn{1}{c|}{\multirow{2}{*}{0.5}} & 20 & 1.712 & 5.009 & $\bm{0.487}(3)$ & 1.567 & 3.170 & $\bm{0.360}(4)$ & 1.836 & 16.667 & $\bm{1.375}(6)$ \\
\multicolumn{1}{c|}{} & 5 & 1.667 & 2.426 & $\bm{0.230}(2)$ & 1.334 & 2.427 & $\bm{0.275}(2)$ & 1.841 & 7.298 & $\bm{1.130}(5)$ \\ 
\Xhline{3\arrayrulewidth}
\end{tabular}}}
\end{sc}
\label{tab:times_data}
\end{table}
We benchmark \texttt{SsNAL-EN} against two different versions of the \emph{coordinate descent} algorithm, one implemented in the \texttt{python package sklearn} and one implemented in the \texttt{R package glmnet} which is written in \texttt{fortran}.
We tested other algorithms, such as \emph{ADMM} and the \emph{proximal gradient} algorithm.
The computational burden of these algorithms for the Elastic Net, just like for the Lasso, is more than two orders of magnitude larger than the one of our approach -- see \citet{li2018} (we are not reporting results here since they could not even complete most of the instances). 

Our simulated data is generated as follows. The entries of the design matrix $A \in \mathbb{R}^{n+m}$ are drawn from a standard normal distribution. We compute the response vector as $b = Ax_t + \epsilon$, where $x_t \in \mathbb{R}^n$ is a sparse vector with $n_0$ non-zeros values all equal to $x^*=5$, and $\epsilon_i \sim N(0, \text{s}_{\epsilon})$ are error terms. 
We fix $\text{s}_{\epsilon}$ as to have a \emph{signal to noise ratio} snr $= \text{var} (Ax_t)/ \text{s}_{\epsilon}^2 = 5$.
\texttt{SsNAL-EN} is run with the tolerance fixed at 1e-6, and $\mu$ in \eqref{eq:line_search} set to 0.2. We start from $\sa^0 = 5$e-3 and increase it by a factor of 5 every iteration. If we start from smaller values of $\sa$, the algorithm needs more iterations to converge, while if $\sa^0$ is too large, \texttt{SsNAL-EN} does not converge to the optimal solution.
We consider three different scenarios characterized by the following values of $(m, n_0, \alpha)$ \\

\centerline{
$ \textbf{sim1:} ~ (500,100,0.6) \qquad
         \textbf{sim2:} ~ (500,20,0.75) \qquad
         \textbf{sim3:} ~ (500,5,0.9)$ 
} 
%
%

We set $\la_1 = \alpha c_\la \la^{max}$, $\la_2 = (1 - \alpha) c_\la \la^{max}$, where $c_\la \in (0,1]$, $\alpha \in [0,1]$, and $\la^{max} = \lVert A^Tb \rVert_{\infty} / \alpha$.
Note
that for \texttt{glmnet} and \texttt{sklearn} we need to divide $\la^{max}$ by $m$ since in the objective function \eqref{eq:objective_function} they divide the square loss for the number of observations. 
\textbf{sim1}, \textbf{sim2} and \textbf{sim3} present an increasing level of sparsity. The sparser is the problem, the larger is the weight we attribute to $\la_1$ relative to $\la_2$ with the different choices of $\alpha$. 

Table \ref{tab:times} reports the CPU time of \texttt{SsNAL-EN}, \texttt{sklearn} and \texttt{glment}.  For each scenario, we consider different values of $n$ and we select the largest $c_\la$ which gives a solution with $n_0$ active components. \texttt{SsNAL-EN} is the fastest algorithm in every instance. The gain with respect to other algorithms increases for larger $n$'s and sparser scenarios, where \texttt{SsNAL-EN} is between 20 and 30 times faster than \texttt{glmnet} and more than 60 times faster than \texttt{sklearn}. 
%
%

We also test all the solvers on some widely studied reference data sets from the \href{https://www.csie.ntu.edu.tw/~cjlin/libsvmtools/datasets/regression.html}{LIBSVM library} \citep{chang2011libsvm}. In particular, we consider \textbf{housing8, bodyfat8,} and \textbf{traizines4}.
For each data set, we create a very large number of features including all terms in a polynomial basis expansion \citep{huang2010polynom}. 
The number reported after the name indicates the order of the expansion. 
The terms in the polynomial expansions are highly collinear, making these examples suitable for the Elastic Net.
To gauge the level of collinearity we compute the largest eigenvalue of $AA^T$ and we normalize it by the number of features $n$. We indicate this number as $\hat \rho$. 
Note how $\hat \rho$ is close to 1 for all the simulated data sets (Table~\ref{tab:times}) and much larger for the data sets created with the polynomial expansions (Table~\ref{tab:times_data}).
Table \ref{tab:times_data} also reports the CPU time of \texttt{SsNAL-EN}, \texttt{sklearn} and \texttt{glment} for the latter. For each algorithm we select the two values of $c_{\la}$ which give an active set of cardinality $20$ and $5$, respectively.
\texttt{SsNAL-EN} is again the fastest algorithm in every instance.  For higher values of $\alpha$ and $r$, it is more than $50$ times faster than \texttt{sklearn} -- the other python-coded algorithm. Furthermore, unlike \texttt{sklearn}, the performance of \texttt{SsNAL-EN} is not affected by the choice of $\alpha$.
Finally, Tables \ref{tab:times} and \ref{tab:times_data} 
report the total number of iterations needed by \texttt{SsNAL-EN} to converge. In all cases convergence is reached in no more than $6$ iteration. 
Notably, $\alpha$ does affect convergence; if we decrease its value, giving more weigh to the $l_2$ norm penalty, convergence is generally reached with just $2$ iterations.
We investigated prediction performance. Results are not reported since the three methods solves the same objective function and converge to the same solution.

Additional results are described in the Supplementary Material.
In Supplement \ref{app:sim_standard_errors}, we report computing time \emph{standard errors} over 20 replications of the same scenario based on {\bf sim1}.
Results confirm the competitiveness of our method. 
\texttt{SsNAL-EN} has the lowest mean computing time and comparable standard errors with respect to \texttt{sklearn} and \texttt{glmnet}.
In Supplement \ref{app:sim_diff_values}, we investigate different values of $m$, $snr$, $\alpha$ and $b$. 
The relative gain of \texttt{SsNAL-EN} with respect to \texttt{sklearn} is even larger for bigger values of $m$ and smaller values of $\alpha$. 
In Supplement \ref{app:sim_screening_solvers}, we benchmark \texttt{SsNAL-EN} against three advanced solvers which implement screening rules: the \texttt{R package biglasso} (written in \texttt{C}$++$), and the \texttt{python packages Gap Safe Rules (GSR)}  and \texttt{celer}. 
\texttt{SsNAL-EN} is faster than all the competitors in very sparse scenarios ($r \approx 10$ selected features). 
In intermediate scenarios ($r \approx 100$) \texttt{biglasso} and \texttt{SsNAL-EN} are comparable and slightly faster than the other algorithms. 
In non-sparse scenarios ($r > 300$) \texttt{biglasso} and \texttt{celer} are about 2 times faster than all other solvers -- as expected, in this case, \texttt{SsNAL-EN} cannot exploit sparsity and looses part of its efficiency. 
Finally, in Supplement \ref{app:sim_solution_path}, we report computing time for a solution path where multiple values of $c_\la$ are considered. We tested \texttt{SsNAL-EN}, texttt{sklearn}, \texttt{glmnet} and \texttt{biglasso}, which have a solution path implementation for $\alpha \ne 1$. Again, \texttt{SsNAL-EN} outperforms the other solvers in almost every instance, being at least 10 times faster than texttt{sklearn}.
The results reported in the Supplement provide further evidence in support of our method -- also considering that some competitors, such as \texttt{glmnet} and \texttt{biglasso}, are highly optimized (in particular for a solution path search), and many use screening, which increases speed but may not find the global minimizer.

\subsection{INSIGHT application}
\label{subsec:insight_application}

\begin{table}[]
\caption{SNPs selected by \emph{SsSNAL-EN} for the CWG and BMI regressions (INSIGHT data). The active sets comprise 13 and 6 SNPs, respectively, and the $\hat x$ are their estimated coefficients. SNPs in red are those selected with an active set of 1. Chromosomes and genes associated to the SNPs were obtained from the \href{https://www.ncbi.nlm.nih.gov/search/}{U.S. National Library of Medicine}. 
}
\small
\centerline{
\scalebox{0.9}{
\begin{tabular}{r|rrc|r|rrc|rrrc}
\Xhline{3\arrayrulewidth}
\multicolumn{8}{c|}{\textbf{CWG}} & \multicolumn{4}{c}{\textbf{BMI}} \\ 
\Xhline{3\arrayrulewidth}
\multicolumn{1}{c|}{\textit{\textbf{snp}}} & \multicolumn{1}{c}{$\bm{\hat x}$} & \multicolumn{1}{c}{\textit{\textbf{chr}}} & \textit{\textbf{gene}} & \multicolumn{1}{c|}{\textit{\textbf{snp}}} & \multicolumn{1}{c}{$\bm{\hat x}$} & \multicolumn{1}{c}{\textit{\textbf{chr}}} & \textit{\textbf{gene}} & \multicolumn{1}{c|}{\textit{\textbf{snp}}} & \multicolumn{1}{c}{$\bm{\hat x}$} & \multicolumn{1}{c}{\textit{\textbf{chr}}} & \textit{\textbf{gene}} \\ 
\Xhline{3\arrayrulewidth}
rs4574484 & 0.19 & 4 & JAKMIP1 & rs60032759 & -0.17 & 4 & - & \multicolumn{1}{r|}{rs77799452} & 0.23 & 1 & IFFO2 \\
rs62295044 & 0.06 & 4 & - & {\red $\bm{rs7697195}$} & 0.22 & 4 & - & \multicolumn{1}{r|}{rs17047916} & -0.25 & 2 & - \\
rs66827781 & 0.06 & 4 & - & rs58516574 & 0.22 & 10 & ARID5B & \multicolumn{1}{r|}{rs10074748} & 0.30 & 5 & - \\
rs139058293 & 0.06 & 4 & - & rs4353245 & -0.06 & 11 & - & \multicolumn{1}{r|}{rs10822135} & -0.27 & 10 & - \\
rs115181650 & 0.07 & 4 & - & rs1893921 & -0.06 & 11 & - & \multicolumn{1}{r|}{rs184636920} & 0.24 & 10 & RNLS \\
rs200024438 & 0.07 & 4 & - & rs12577363 & -0.06 & 11 & - & \multicolumn{1}{r|}{{\red $\bm{rs79187646}$}} & 0.32 & 11 & {\bf {\red NTM}} \\
rs78918827 & 0.07 & 4 & - &  &  &  &  & \multicolumn{1}{l|}{} & \multicolumn{1}{l}{} & \multicolumn{1}{l}{} & \multicolumn{1}{l}{} \\ 
\Xhline{3\arrayrulewidth}
\end{tabular}}}
\label{tab:cwg_genes}
\vspace{-0.4cm}
\end{table}
Here, we apply \texttt{SsNAL-EN} to data from the Intervention Nurses Start Infants Growing on Healthy Trajectories (INSIGHT) study \citep{paul2014intervention}. 
One goal of INSIGHT is to investigate genetic variants that affect the risk of childhood obesity. 
In particular, we look for relevant Single Nucleotide Polymorphisms (SNPs). SNPs have been recently related to obesity phenotypes by several Genome-Wide Association Studies (GWASs), e.g~\cite{locke2015genetic}. We analyze
two different outcome measurements: \emph{Conditional Weight Gain} (CWG), which describes the change in weight between birth and six months \citep{taveras2009weight}, and \emph{Body Mass Index} at age 3 (BMI).
After 
preprocessing,
the design matrix for CWG 
comprises $342594$ SNPs and $226$ observations, and that for BMI $342325$ SNPs and $210$ observations.

Figure \ref{fig:cwg_tuning} displays the parameter tuning criteria for both responses. 
All three criteria considered identify just {\em one} dominant SNP.
\emph{gcv} and \emph{e-bic} present a second elbow corresponding to an active set of 13 SNPs for CWG, and 6 SNPs for BMI. 
SNPs and estimated coefficients are reported in Table~\ref{tab:cwg_genes}.
The table also contains the \emph{chromosome}, the \emph{position} and the \emph{gene} associated with a SNP (when available) according to the \href{https://www.ncbi.nlm.nih.gov/search/}{U.S. National Library of Medicine} website. 
In depth biological interpretations of these results 
are beyond the scope of the present manuscript. However, we can still highlight some facts. Both responses point towards very parsimonious models and the two active sets of 13 and 6 SNPs do not overlap -- suggesting that, despite their sizeable correlation (0.545), CWG and BMI may be at least partially driven by different mechanisms.
The two top SNPs for CWG ($rs7697195$) and BMI ($rs79187646$) are different and located on different chromosomes (4 and 11, respectively), and they do not appear to co-occur on INSIGHT individuals (their correlation is only 0.004). 
While we could not link $rs7697195$ to any gene, $rs79187646$ -- the dominant SNP for BMI -- is located in the well known NTM gene.
According to the \href{https://www.ebi.ac.uk/gwas/home}{NHGRI-EBI GWAS Catalog}, NTM has been identified in a wide range of GWASs connecting it to BMI, food addiction, and other obesity-related traits. e.g~\citep{kichaev2019bmi2, pulit2019bmi1}. 
Finally, we note that also $rs58516574$, which is the SNP with the second largest estimated coefficient for CWG, is located in a known gene; ARID5B. Again according to \href{https://www.ebi.ac.uk/gwas/home}{NHGRI-EBI GWAS Catalog},
ARID5B is associated to various traits, including some which could be related to obesity (e.g., waste to hip ratio).

%
%

\section{Conclusions and future work}
\label{sec:conclusions}
\vspace{-0.1cm}

We developed a new efficient Semi-smooth Newton method to solve the Elastic Net problem in high-dimensional settings. This method wisely exploits 
the sparsity induced by the penalty and the sparsity inherent to the augmented Lagrangian problem. 
We provided effective algorithms to solve both the augmented Lagrangian problem and the inner-subproblem, proving the super-linear convergence of both.
Simulation results based on synthetic and real data show a very large gain in computational efficiency with respect to the best existing algorithms.
We also used our method in a GWAS application, identifying genetic variants associated with childhood obesity. 

In the near future, we plan to adapt \texttt{SsNAL-EN} to the function-on-scalar regression setting. This framework is particularly relevant for genetic studies where very large number of SNPs are regressed against 
outcomes suitable for Functional Data Analysis \citep{cremona2019functional}. For instance, in addition to CWG and BMI, the INSIGHT data contains also longitudinal growth information on children \citep{craig2019polygenic}. A functional version of \texttt{SsNAL-EN} will involve more complex penalties and build upon the results presented in this article.

%
%

\section*{Broader Impact}

We present a new, mathematically sound, practically interpretable and computationally very efficient approach to perform selection of relevant features in very high-dimensional problems. With the advent of big data, the proposed methodology may have a concrete and substantial impact in a wide range of scientific fields and applications -- including (but by no means limited to) biomedical research. We have presented an application to a Genome-Wide Association Studies -- such studies, investigating association between complex human diseases and genetic variants, are now ubiquitous and ever growing  in size. Our method is perfectly suited to aid in the detection of genetic risk factors underlying important health pathologies. From a technical point of view, our work straddles statistics, optimization and computer science -- integrating multiple STEM components and paving the way to the development of yet more sophisticated methodologies. Driven by scientific questions based on the use of complex disease phenotypes as outcomes in GWAS, we plan to extend our approach beyond the Elastic Net to tackle feature selection in function-on-scalar regression problems. 


%
%

\begin{ack}
This work was partially fund by NSF DMS-1712826 and the Huck Institutes of Life Sciences at Penn State. The authors thank Kateryna Makova  for sharing the INSIGHT data, Sarah Craig and Ana Maria Kenney for the fruitful discussions on the data, and Ludovica Delpopolo for her valuable help in the algorithm implementation. 
\end{ack}

%
%


\bibliographystyle{chicago}

\begin{thebibliography}{}

\bibitem[\protect\citeauthoryear{Beck and Teboulle}{Beck and
  Teboulle}{2009}]{beck2009fast}
Beck, A. and M.~Teboulle (2009).
\newblock A fast iterative shrinkage-thresholding algorithm for linear inverse
  problems.
\newblock {\em SIAM journal on imaging sciences\/}~{\em 2\/}(1), 183--202.

\bibitem[\protect\citeauthoryear{Belloni, Chernozhukov, and Hansen}{Belloni
  et~al.}{2014}]{belloni2014debiasedols}
Belloni, A., V.~Chernozhukov, and C.~Hansen (2014).
\newblock Inference on treatment effects after selection among high-dimensional
  controls.
\newblock {\em The Review of Economic Studies\/}~{\em 81\/}(2), 608--650.

\bibitem[\protect\citeauthoryear{Boyd, Parikh, Chu, Peleato, Eckstein,
  et~al.}{Boyd et~al.}{2011}]{boyd2011distributed}
Boyd, S., N.~Parikh, E.~Chu, B.~Peleato, J.~Eckstein, et~al. (2011).
\newblock Distributed optimization and statistical learning via the alternating
  direction method of multipliers.
\newblock {\em Foundations and Trends{\textregistered} in Machine
  learning\/}~{\em 3\/}(1), 1--122.

\bibitem[\protect\citeauthoryear{Boyd and Vandenberghe}{Boyd and
  Vandenberghe}{2004}]{boyd2004convex}
Boyd, S. and L.~Vandenberghe (2004).
\newblock {\em Convex optimization}.
\newblock Cambridge university press.

\bibitem[\protect\citeauthoryear{Chang and Lin}{Chang and
  Lin}{2011}]{chang2011libsvm}
Chang, C.-C. and C.-J. Lin (2011).
\newblock Libsvm: A library for support vector machines.
\newblock {\em ACM transactions on intelligent systems and technology
  (TIST)\/}~{\em 2\/}(3), 1--27.

\bibitem[\protect\citeauthoryear{Chen and Chen}{Chen and
  Chen}{2012}]{chen2012extended}
Chen, J. and Z.~Chen (2012).
\newblock Extended bic for small-n-large-p sparse glm.
\newblock {\em Statistica Sinica\/}, 555--574.

\bibitem[\protect\citeauthoryear{Clarke}{Clarke}{1990}]{clarke1990optimization}
Clarke, F.~H. (1990).
\newblock {\em Optimization and nonsmooth analysis}, Volume~5.
\newblock Siam.

\bibitem[\protect\citeauthoryear{Correa, Jofre, and Thibault}{Correa
  et~al.}{1992}]{correa1992subcharacterization}
Correa, R., A.~Jofre, and L.~Thibault (1992).
\newblock Characterization of lower semicontinuous convex functions.
\newblock {\em Proceedings of the American Mathematical Society\/}, 67--72.

\bibitem[\protect\citeauthoryear{Craig, Kenney, Lin, Paul, Birch, Savage,
  Marini, Chiaromonte, Reimherr, and Makova}{Craig
  et~al.}{2019}]{craig2019polygenic}
Craig, S. J.~C., A.~M. Kenney, J.~Lin, I.~M. Paul, L.~L. Birch, J.~Savage,
  M.~E. Marini, F.~Chiaromonte, M.~L. Reimherr, and K.~D. Makova (2019).
\newblock Polygenic risk score based on weight gain trajectories is a strong
  predictor of childhood obesity.
\newblock {\em bioRxiv\/}, 606277.

\bibitem[\protect\citeauthoryear{Cremona, Xu, Makova, Reimherr, Chiaromonte,
  and Madrigal}{Cremona et~al.}{2019}]{cremona2019functional}
Cremona, M.~A., H.~Xu, K.~D. Makova, M.~Reimherr, F.~Chiaromonte, and
  P.~Madrigal (2019).
\newblock Functional data analysis for computational biology.
\newblock {\em Bioinformatics (Oxford, England)\/}~{\em 35\/}(17), 3211.

\bibitem[\protect\citeauthoryear{Deng and So}{Deng and So}{2019}]{deng2019}
Deng, Z. and A.~M.-C. So (2019).
\newblock An efficient augmented lagrangian based method for constrained lasso.
\newblock {\em arXiv preprint arXiv:1903.05006\/}.

\bibitem[\protect\citeauthoryear{Dontchev and Rockafellar}{Dontchev and
  Rockafellar}{2009}]{dontchev2009implicit}
Dontchev, A.~L. and R.~T. Rockafellar (2009).
\newblock Implicit functions and solution mappings.
\newblock {\em Springer Monographs in Mathematics. Springer\/}~{\em 208}.

\bibitem[\protect\citeauthoryear{D{\"u}nner, Forte, Tak{\'a}{\v{c}}, and
  Jaggi}{D{\"u}nner et~al.}{2016}]{dunner2016primal}
D{\"u}nner, C., S.~Forte, M.~Tak{\'a}{\v{c}}, and M.~Jaggi (2016).
\newblock Primal-dual rates and certificates.
\newblock {\em arXiv preprint arXiv:1602.05205\/}.

\bibitem[\protect\citeauthoryear{Fan and Li}{Fan and
  Li}{2001}]{fan2001variable}
Fan, J. and R.~Li (2001).
\newblock Variable selection via nonconcave penalized likelihood and its oracle
  properties.
\newblock {\em Journal of the American statistical Association\/}~{\em
  96\/}(456), 1348--1360.

\bibitem[\protect\citeauthoryear{Fenchel}{Fenchel}{1949}]{fenchel1949conjugate}
Fenchel, W. (1949).
\newblock On conjugate convex functions.
\newblock {\em Canadian Journal of Mathematics\/}~{\em 1\/}(1), 73--77.

\bibitem[\protect\citeauthoryear{Friedman, Hastie, and Tibshirani}{Friedman
  et~al.}{2010}]{friedman2010regularization}
Friedman, J., T.~Hastie, and R.~Tibshirani (2010).
\newblock Regularization paths for generalized linear models via coordinate
  descent.
\newblock {\em Journal of statistical software\/}~{\em 33\/}(1), 1.

\bibitem[\protect\citeauthoryear{Gaines, Kim, and Zhou}{Gaines
  et~al.}{2018}]{gaines2018constrained}
Gaines, B.~R., J.~Kim, and H.~Zhou (2018).
\newblock Algorithms for fitting the constrained lasso.
\newblock {\em Journal of Computational and Graphical Statistics\/}~{\em
  27\/}(4), 861--871.

\bibitem[\protect\citeauthoryear{Hiriart-Urruty, Strodiot, and
  Nguyen}{Hiriart-Urruty et~al.}{1984}]{hiriart1984hessian}
Hiriart-Urruty, J.-B., J.-J. Strodiot, and V.~H. Nguyen (1984).
\newblock Generalized hessian matrix and second-order optimality conditions for
  problems withc 1, 1 data.
\newblock {\em Applied mathematics and optimization\/}~{\em 11\/}(1), 43--56.

\bibitem[\protect\citeauthoryear{Huang, Jia, Yu, Chun, Maniatis, and
  Naik}{Huang et~al.}{2010}]{huang2010polynom}
Huang, L., J.~Jia, B.~Yu, B.-G. Chun, P.~Maniatis, and M.~Naik (2010).
\newblock Predicting execution time of computer programs using sparse
  polynomial regression.
\newblock In {\em Advances in neural information processing systems}, pp.\
  883--891.

\bibitem[\protect\citeauthoryear{Jansen}{Jansen}{2015}]{jansen2015generalized}
Jansen, M. (2015).
\newblock Generalized cross validation in variable selection with and without
  shrinkage.
\newblock {\em Journal of statistical planning and inference\/}~{\em 159},
  90--104.

\bibitem[\protect\citeauthoryear{Kichaev, Bhatia, Loh, Gazal, Burch, Freund,
  Schoech, Pasaniuc, and Price}{Kichaev et~al.}{2019}]{kichaev2019bmi2}
Kichaev, G., G.~Bhatia, P.-R. Loh, S.~Gazal, K.~Burch, M.~K. Freund,
  A.~Schoech, B.~Pasaniuc, and A.~L. Price (2019).
\newblock Leveraging polygenic functional enrichment to improve gwas power.
\newblock {\em The American Journal of Human Genetics\/}~{\em 104\/}(1),
  65--75.

\bibitem[\protect\citeauthoryear{Li and Lin}{Li and Lin}{2015}]{li2015proxgrad}
Li, H. and Z.~Lin (2015).
\newblock Accelerated proximal gradient methods for nonconvex programming.
\newblock In {\em Advances in neural information processing systems}, pp.\
  379--387.

\bibitem[\protect\citeauthoryear{Li, Sun, and Toh}{Li et~al.}{2018}]{li2018}
Li, X., D.~Sun, and K.-C. Toh (2018).
\newblock A highly efficient semismooth newton augmented lagrangian method for
  solving lasso problems.
\newblock {\em SIAM Journal on Optimization\/}~{\em 28\/}(1), 433--458.

\bibitem[\protect\citeauthoryear{Locke, Kahali, Berndt, Justice, Pers, Day,
  Powell, Vedantam, Buchkovich, Yang, et~al.}{Locke
  et~al.}{2015}]{locke2015genetic}
Locke, A.~E., B.~Kahali, S.~I. Berndt, A.~E. Justice, T.~H. Pers, F.~R. Day,
  C.~Powell, S.~Vedantam, M.~L. Buchkovich, J.~Yang, et~al. (2015).
\newblock Genetic studies of body mass index yield new insights for obesity
  biology.
\newblock {\em Nature\/}~{\em 518\/}(7538), 197.

\bibitem[\protect\citeauthoryear{Luque}{Luque}{1984}]{luque1984asymptotic}
Luque, F.~J. (1984).
\newblock Asymptotic convergence analysis of the proximal point algorithm.
\newblock {\em SIAM Journal on Control and Optimization\/}~{\em 22\/}(2),
  277--293.

\bibitem[\protect\citeauthoryear{Massias, Gramfort, and Salmon}{Massias
  et~al.}{2018}]{massias2018celer}
Massias, M., A.~Gramfort, and J.~Salmon (2018).
\newblock Celer: a fast solver for the lasso with dual extrapolation.
\newblock {\em arXiv preprint arXiv:1802.07481\/}.

\bibitem[\protect\citeauthoryear{Ndiaye, Fercoq, Gramfort, and Salmon}{Ndiaye
  et~al.}{2017}]{ndiaye2017gap}
Ndiaye, E., O.~Fercoq, A.~Gramfort, and J.~Salmon (2017).
\newblock Gap safe screening rules for sparsity enforcing penalties.
\newblock {\em The Journal of Machine Learning Research\/}~{\em 18\/}(1),
  4671--4703.

\bibitem[\protect\citeauthoryear{Parikh, Boyd, et~al.}{Parikh
  et~al.}{2014}]{parikh2014proximal}
Parikh, N., S.~Boyd, et~al. (2014).
\newblock Proximal algorithms.
\newblock {\em Foundations and Trends{\textregistered} in Optimization\/}~{\em
  1\/}(3), 127--239.

\bibitem[\protect\citeauthoryear{Paul, Williams, Anzman-Frasca, Beiler, Makova,
  Marini, Hess, Rzucidlo, Verdiglione, Mindell, et~al.}{Paul
  et~al.}{2014}]{paul2014intervention}
Paul, I.~M., J.~S. Williams, S.~Anzman-Frasca, J.~S. Beiler, K.~D. Makova,
  M.~E. Marini, L.~B. Hess, S.~E. Rzucidlo, N.~Verdiglione, J.~A. Mindell,
  et~al. (2014).
\newblock The intervention nurses start infants growing on healthy trajectories
  (insight) study.
\newblock {\em BMC pediatrics\/}~{\em 14\/}(1), 184.

\bibitem[\protect\citeauthoryear{Pulit, Stoneman, Morris, Wood, Glastonbury,
  Tyrrell, Yengo, Ferreira, Marouli, Ji, et~al.}{Pulit
  et~al.}{2019}]{pulit2019bmi1}
Pulit, S.~L., C.~Stoneman, A.~P. Morris, A.~R. Wood, C.~A. Glastonbury,
  J.~Tyrrell, L.~Yengo, T.~Ferreira, E.~Marouli, Y.~Ji, et~al. (2019).
\newblock Meta-analysis of genome-wide association studies for body fat
  distribution in 694 649 individuals of european ancestry.
\newblock {\em Human molecular genetics\/}~{\em 28\/}(1), 166--174.

\bibitem[\protect\citeauthoryear{Robinson}{Robinson}{1981}]{robinson1981some}
Robinson, S.~M. (1981).
\newblock Some continuity properties of polyhedral multifunctions.
\newblock In {\em Mathematical Programming at Oberwolfach}, pp.\  206--214.
  Springer.

\bibitem[\protect\citeauthoryear{Rockafellar}{Rockafellar}{1976a}]{rockafellar1976augmented}
Rockafellar, R.~T. (1976a).
\newblock Augmented lagrangians and applications of the proximal point
  algorithm in convex programming.
\newblock {\em Mathematics of operations research\/}~{\em 1\/}(2), 97--116.

\bibitem[\protect\citeauthoryear{Rockafellar}{Rockafellar}{1976b}]{rockafellar1976}
Rockafellar, R.~T. (1976b).
\newblock Monotone operators and the proximal point algorithm.
\newblock {\em SIAM journal on control and optimization\/}~{\em 14\/}(5),
  877--898.

\bibitem[\protect\citeauthoryear{Rockafellar and Wets}{Rockafellar and
  Wets}{2009}]{rockafellar2009variational}
Rockafellar, R.~T. and R.~J.-B. Wets (2009).
\newblock {\em Variational analysis}, Volume 317.
\newblock Springer Science \& Business Media.

\bibitem[\protect\citeauthoryear{Taveras, Rifas-Shiman, Belfort, Kleinman,
  Oken, and Gillman}{Taveras et~al.}{2009}]{taveras2009weight}
Taveras, E.~M., S.~L. Rifas-Shiman, M.~B. Belfort, K.~P. Kleinman, E.~Oken, and
  M.~W. Gillman (2009).
\newblock Weight status in the first 6 months of life and obesity at 3 years of
  age.
\newblock {\em Pediatrics\/}~{\em 123\/}(4), 1177--1183.

\bibitem[\protect\citeauthoryear{Tibshirani}{Tibshirani}{1996}]{tibshirani1996regression}
Tibshirani, R. (1996).
\newblock Regression shrinkage and selection via the lasso.
\newblock {\em Journal of the Royal Statistical Society: Series B
  (Methodological)\/}~{\em 58\/}(1), 267--288.

\bibitem[\protect\citeauthoryear{Tibshirani, Taylor, et~al.}{Tibshirani
  et~al.}{2012}]{tibshirani2012degrees}
Tibshirani, R.~J., J.~Taylor, et~al. (2012).
\newblock Degrees of freedom in lasso problems.
\newblock {\em The Annals of Statistics\/}~{\em 40\/}(2), 1198--1232.

\bibitem[\protect\citeauthoryear{Touchette}{Touchette}{2005}]{touchette2005legendre}
Touchette, H. (2005).
\newblock Legendre-fenchel transforms in a nutshell.
\newblock {\em URL http://www. maths. qmul. ac. uk/\~{} ht/archive/lfth2.
  pdf\/}.

\bibitem[\protect\citeauthoryear{Tseng and Yun}{Tseng and
  Yun}{2009}]{tseng2009coordinate}
Tseng, P. and S.~Yun (2009).
\newblock A coordinate gradient descent method for nonsmooth separable
  minimization.
\newblock {\em Mathematical Programming\/}~{\em 117\/}(1-2), 387--423.

\bibitem[\protect\citeauthoryear{Zhao, Shojaie, and Witten}{Zhao
  et~al.}{2017}]{zhao2017debiasedols}
Zhao, S., A.~Shojaie, and D.~Witten (2017).
\newblock In defense of the indefensible: A very naive approach to
  high-dimensional inference.
\newblock {\em arXiv preprint arXiv:1705.05543\/}.

\bibitem[\protect\citeauthoryear{Zhao, Sun, and Toh}{Zhao
  et~al.}{2010}]{zhao2010newton}
Zhao, X.-Y., D.~Sun, and K.-C. Toh (2010).
\newblock A newton-cg augmented lagrangian method for semidefinite programming.
\newblock {\em SIAM Journal on Optimization\/}~{\em 20\/}(4), 1737--1765.

\bibitem[\protect\citeauthoryear{Zhou and So}{Zhou and
  So}{2017}]{zhou2017unified}
Zhou, Z. and A.~M.-C. So (2017).
\newblock A unified approach to error bounds for structured convex optimization
  problems.
\newblock {\em Mathematical Programming\/}~{\em 165\/}(2), 689--728.

\bibitem[\protect\citeauthoryear{Zou}{Zou}{2006}]{zou2006adaptive}
Zou, H. (2006).
\newblock The adaptive lasso and its oracle properties.
\newblock {\em Journal of the American statistical association\/}~{\em
  101\/}(476), 1418--1429.

\bibitem[\protect\citeauthoryear{Zou and Hastie}{Zou and
  Hastie}{2005}]{zou2005regularization}
Zou, H. and T.~Hastie (2005).
\newblock Regularization and variable selection via the elastic net.
\newblock {\em Journal of the royal statistical society: series B (statistical
  methodology)\/}~{\em 67\/}(2), 301--320.

\end{thebibliography}

%
%

\appendix
\counterwithin{figure}{section}
\counterwithin{table}{section}
\counterwithin{equation}{section}

\newpage
\begin{center}
\large
\rule{\textwidth}{0.05cm}
	 $\phantom{i}$ \\
	 {\bf Supplementary Material} \\
	  \small{$\phantom{i}$}
\rule{\textwidth}{0.05cm}
\end{center}

%
%
\section{Proof of Proposition 1}
\label{app:proof_p_star}
\setcounter{equation}{0}

Consider $p(x) = \la_1 \norm{x}_1 + \frac{\la_2}{2} \norm{x}_2$, with $x \in \mathbb{R}^n$. To compute $p^*(z)$, we use the following result \citep{touchette2005legendre}:
\begin{equation}
\small
\label{eq:find_conj}
     	p^*(z) = z^T \bar{x} - p\l(\bar x\r), 
\end{equation}
 \noindent
 where $\bar x = \arg \sup_x P(x, z) = \arg \sup_x \left( z^T x - p(x) \right)$. To find $\bar x$ one has to solve $\nabla_x P(x, z) = 0$, which is equivalent to solve $z = \nabla_x P(x)$ for $x$ given $z$. In our case, we have: 
\begin{equation}
\small
\label{eq:find_conj_our}
	p^*(z) = z^T \bar{x} - \la_1 \norm{\bar x}_1 - \frac{\la_2}{2} \norm{\bar x}_2^2
\end{equation}
\noindent
If we compute $\frac{\partial}{\partial x_i}$ $P(x, z)$, and we set it equal to 0, we obtain:
\begin{equation}
\small
\label{eq:x_bar_i}
	\tilde x_i =
	\begin{cases}
 			(z_i - \la_1) / \la_2, & x_i > 0\\
 			\in \l(z_i - \la_1 \lbrack -1, 1 \rbrack \r) / \la_2, & x_i = 0 \\
 			(z_i + \la_1) / \la_2, & x_i < 0
  	\end{cases},
\end{equation}
\noindent 
where $\lbrack -1, 1 \rbrack = \partial \norm {x_i}_1$ at $x_i = 0$.  To find $\bar x$, we need to take into account the fact that $\text{dom}(p) = \text{range}\l(p^*\r)$ \citep{touchette2005legendre} and transform $\tilde x_i$ consequently. The $i$-th component of $\bar x$ is given by: 
\begin{equation}
\small
\label{eq:s_i}
	\bar x_i =
	\begin{cases}
 			(z_i - \la_1) / \la_2, & z_i \ge \la_1 \\
 			0, & |z_i| < \la_1 \\
 			(z_i + \la_1)/ \la_2, & z_i \le -\la_1
  	\end{cases}.
\end{equation}
\noindent 
By \eqref{eq:find_conj_our}, we can compute the $i$-th component of $p^*(z)$ as $z_i \bar{x_i} - \la_1 |\bar x_i| - (\la_2/2) \bar x_i^2$. We have:
\begin{equation}
\small
\label{eq:p_star_i}
	\begin{split}
	 	p^*\l(z_i\r) &= 
	 		\begin{cases}
 				z_i(z_i - \la_1)/\la_2 - \la_1(z_i - \la_1)/\la_2  - \l(\la_2/2\r) (z_i - \la_1)^2/\la_2^2  , & z_i \ge \la_1 \\
 				0, & |z_i| < \la_1 \\
 				z_i(z_i + \la_1)/\la_2 - \la_1(z_i + \la_1)/\la_2 - \l(\la_2/2\r) (z_i - \la_1)^2/\la_2^2  , & z_i \le \- \la_1
 			\end{cases} = \\
  			&= \frac{1}{2 \la_2} 
  				\begin{cases}
  					(z_i - \la_1)^2, & z_i \ge \la_1 \\
  					0, &  |z_i| < \la_1\\
 					(z_i + \la_1)^2, & z_i \le \la_1 
  				\end{cases}
  		\end{split}
\end{equation}
\noindent
Now, we just need to show:
\begin{equation}
\small
\label{eq:need_to_show_p_star}
	z^T \bar{x} - \la_1 \norm{\bar x}_1 - \frac{\la_2}{2} \norm{\bar x}_2^2 = \sum_{i=1}^n p^*\l(z_i\r).
\end{equation}
\noindent
But, we have:
\begin{align}
\small
\label{eq:final_step_p_star}
	\begin{split}
			z^T \bar{x} - \la_1 \norm{\bar x}_1 - \frac{\la_2}{2} \norm{\bar x}_2^2 &=
		    \sum_{i=1}^n z_i \bar x_i - 
		    \la_1 \sum_{i=1}^n \la_1 |\bar x_i| - \frac{\la_2}{2} \sum_{i=1}^n \bar x_i^2 = \\
			&=  \sum_{i=1}^n{ \l( z_i \bar{x_i} - \la_1 |\bar x_i| - (\la_2/2) \bar x_i^2\r)} = \\
			& = \sum_{i=1}^n p^*\l(z_i\r).
 	\end{split}
\end{align}

%
%
\newpage
\section{Proof of Proposition 2}
\label{app:proof_psi}
\setcounter{equation}{0}

\subsubsection*{Part 2}
We start proving the second part of the proposition, i.e:
\begin{equation}
\small
\label{eq:z_bar}
	\bar z =  \prox_{p^*/\sa} \left( x/\sa - A^T \bar y \right).
\end{equation}
\noindent
If we take the derivative with respect to $z$ of $\mathcal{L}_\sa \left(z ~|~ \bar y, x^k \right) 
$ and we set it equal to $0$, we get:
\begin{equation}
\small
\label{eq:der_z_bar}
	\frac{x}{\sa} - A^T \bar y - z = \frac{\nabla p^*(z)}{\sa}.
\end{equation}
\noindent
By the \emph{sub-gradient} characterization of the proximal operators \citep{correa1992subcharacterization}, we know: 
\begin{equation}
\small
\label{eq:sub_char}
	u = \prox_f(t) ~ \text{ if and only if } ~ t - u \in \partial f(u)
\end{equation}
\noindent
Set $t = x / \sa - A^T \bar y$, $u = z$, and $f = p^*/\sa$. The second part of \eqref{eq:sub_char} is true by \eqref{eq:der_z_bar} -- in our case $p^*/\sa$ is differentiable so we have strictly equal to. The first term of \eqref{eq:sub_char}, gives us \eqref{eq:z_bar}.

\subsubsection*{Part 1}

 For the first part of the preposition, we have to prove:
\begin{equation}
\small
\label{eq:psi_y}
	\psi(y) = h^*(y) + \frac{1 + \sa \la_2}{2 \sa} \norm{\prox_{\sa p}\left( x -  \sa A^T y\right) }_2^2 - \frac{1}{2\sa} \norm{x}_2^2.
\end{equation}
\noindent
Note, by Moreau decomposition, we have: $\bar z = x / \sa - A^T y - (1/ \sa) \prox_{\sa p}\l(x - \sa A^T y \r)$. By definition, $\psi(y) = \mathcal{L}_\sa \left(y ~|~ \bar z, x^k \right)$, i.e:
\begin{equation}
\small
\label{eq:psi_y_long}
	\begin{split}
		\psi(y) &= h^*(y) + p^*(\bar z) - x^T\l( A^T y + \bar z \r)	+ \frac{\sa}{2}\norm{A^Ty + \bar z}_2^2 = \\
		 &= h^*(y) + p^*(\bar z) - x^T\l( A^T y + \frac{x}{\sa} - A^T y - \frac{1}{\sa} \prox_{\sa p}\l(x - \sa A^T y\r) \r) + \\
		 &\qquad + \frac{\sa}{2}\norm{A^Ty + \frac{x}{\sa} - A^T y - \frac{1}{\sa} \prox_{\sa p}\l(x - \sa A^T y\r)}_2^2 = \\
		 &= h^*(y) + p^*(\bar z) + \frac{1}{2 \sa} \norm{\prox_{\sa p}\left( x - \sa A^T y\right) }_2^2 - \frac{1}{2\sa} \norm{x}_2^2.
	\end{split}
\end{equation}
\noindent
We need to compute $p^*(\bar z)$, i.e.~$p^*\l( \prox_{p^*/\sa} \left( x/\sa - A^T  y \right)\r)$. In the Lasso \citep{li2015proxgrad} and the constrained Lasso \citep{deng2019}, $p^*$ is an indicator function and $p^*(\bar z)$ is equal to 0. This is not our case. Let define $t = x - \sa A^T y$. Composing the second equation in \eqref{eq:prox_elastic} and \eqref{eq:p_star}, we have:
\begin{equation}
\small
\label{eq:p_star_of_z_bar}
	\begin{split}
		p^*\l( \prox_{p^*/\sa} \left( t / \sa\right)\r) &= \frac{1}{2 \la_2} \sum_{i=1}^n
			\begin{cases}
				\l(\frac{t_i\la_2 + \la_1}{1 + \sa \la_2} - \la_1\r)^2, & t_i \ge \sa \la_1 \\
				0, & |t_i| < \sa \la_1 \\
				\l(\frac{t_i\la_2 - \la_1}{1 + \sa \la_2} + \la_1\r)^2, & t_i \le - \sa \la_1 \\
			\end{cases} = \\
			& = \frac{\la_2}{2} \frac{1}{\l( 1+ \sa \la_2  \r)^2} \sum_{i=1}^n
				\begin{cases}
					\l(t_i - \sa \la_1 \r)^2, & t_i \ge \sa \la_1 \\ 
					0, & |t_i| < \sa \la_1 \\
					\l(t_i + \sa \la_1 \r)^2, & t_i \le - \sa \la_1 \\
				\end{cases} =  \\ 
			&	=\text{by } \eqref{eq:prox_elastic}  = \frac{\la_2}{2} \sum_{i=1}^n \l(\prox_{\sa p}(t_i) \r)^2 
	\end{split}
\end{equation}
\noindent
Therefore, we have $p^*(\bar z) = \frac{\la_2}{2} \norm{\prox_{\sa p}(x -\sa A^T y)}_2^2$. Plugging it in \eqref{eq:psi_y_long}, we prove \eqref{eq:psi_y}.

%
%
\newpage
\section{Convergence Analysis}
\label{app:convergence}
\setcounter{equation}{0}

\subsection{Inexact Augmented Lagrangian Method}

To state the \emph{global convergence} of \textbf{Algorithm \ref{alg:al}} and the \emph{super-linear convergence} of the solution $(y^k, z^k, x^k)$, we refer to \emph{theorem 3.2} and \emph{theorem 3.3} in \citet{li2018} -- which are in turned based on the fundamental results presented in \citet{rockafellar1976augmented, rockafellar1976}, and \citet{luque1984asymptotic}. 
Here, we just need verify that the theorems assumptions hold. In particular, we met the assumptions on $h(\cdot)$, since it is the same for Lasso and Elastic Net, and it is always possible to implement the stopping criteria for the local convergence analysis described in \citet{li2018} -- Section 3. The main challenge is to verify that the operators $\mathcal T_f$ and $\mathcal T_l$ satisfy the \emph{error bound condition}, since they are different from the Lasso case.

Given the closed proper convex function $f$ in the objective  \eqref{eq:objective_function}, and the convex-concave lagrangian function $l$ in \eqref{eq:lagrangian}, we define the maximal monotone operators $\mathcal T_f$ and $\mathcal T_l$ as in \citet{rockafellar1976augmented}:
\begin{equation}
\small
\label{eq:t_operators}
    \mathcal{T}_f(x) = \partial f(x), \quad 
    \mathcal{T}_l (y, z, x) = \{ (y', z', x') | (y', z', -x') \in \partial l(y, z, x)\}.
\end{equation}
We have to show that $\mathcal T_f$ and $\mathcal T_l$ are \emph{metric subregular} \cite{dontchev2009implicit}, or equivalently that they satisfy the \emph{error bound condition} \cite{robinson1981some}, also called \emph{growth condition} \cite{luque1984asymptotic}.

In particular, we say that a multivalue mapping $F~:~\mathcal{X} \rightrightarrows \mathcal{Y}$ satisfies the \emph{error bound condition} at $y \in \mathcal{Y}$ with modulus $\kappa > 0$ if $F^{-1}(y) \ne \emptyset$ and there exists $\epsilon > 0$ such that if $x \in \mathcal{X}$ with $\text{dist}(y, F(x)) \le \epsilon$, then
\begin{equation}
\small
\label{eq:error_bound_cond}
    \text{dist}(x, F^{-1}(y)) \le \kappa \text{dist}(y, F(x)).
\end{equation}
The regularity of $\mathcal{T}_f$ comes from \citet{zhou2017unified}: since $\nabla h$ is  Lipschitz continuous and $p$ has a polyhedral epigraph, $\mathcal{T}_f$ satisfies the error bound condition. 
Verifying the bound condition for $\mathcal{T}_l$ in the Lasso problem is not straightforward. However, in the Elastic Net case, we can use some known results given the special form of $p^*$ in \eqref{eq:p_star}. First, note that $p^*$ is a piecewise linear-quadratic function. Thus, we can apply \emph{Proposition 12.30} in \citet{rockafellar2009variational} and state that the subgradient mapping $\partial p^*$ is piecewise polyhedral. 
Finally, from \cite{robinson1981some} we know that polyhedral multifunctions satisfy the error bound condition for any point $y \in \mathcal{Y}$. This proves the regularity of $\mathcal{T}_l$ and, therefore, the super-linear convergence of the method.

\subsection{Semi-smooth Newton Method}

Again, to state the super-linear convergence of the sequence $\{y^j\}$ produced by the {\emph Semi-smooth Newton Method} in \textbf{Algorithm \ref{alg:al}}, we can use \emph{theorem 3.6} in \citet{li2018}, which is based on the crucial results of \citet{zhao2010newton}. All the assumptions are easy to verify. $\nabla h^*$ and $\prox_{\sa p}$ in \eqref{eq:prox_elastic} are  semi-smooth functions. 
By \emph{proposition 3.3} in \citet{zhao2010newton}, $d^j$ defined in \eqref{eq:newton_direction} is a descent direction. Finally, recall the definition of $V$:
\[
\small
    V := I_m + \sa A Q A^T,
\]
with $Q$ being the diagonal matrix in \eqref{eq:Q_def}. We have $V \in \hat \partial^2 \psi (y)$ and $V$ positive semidefinite.

%
%
\newpage
\section{Simulation Study}
\label{app:sim}

We ran all simulations on a MacBookPro with 3.3 GHz DualCore Intel Core i7 processor and 16GB ram.  We reran all \texttt{python} simulations using \texttt{openblas} and \texttt{mkl} as blas systems, with threads=1,2 and \texttt{openmp}, with threads=1,4. In all scenarios, times match those reported in the paper that are obtained considering \texttt{openblas}  with 2 threads and \texttt{openmp} with 4 threads.

\subsection{Standard errors of computing time for sim1}
\label{app:sim_standard_errors}

\begin{table}[H]
\vspace{-0.5cm}
\caption{\emph{Mean computation time} and \emph{standard error} (in parenthesis) over 20 replications of the same scenario. Data has been generated as described in Section \ref{sec:sim_and_case} and the parameters are set according to \textbf{sim1} (m=500, $n_0$=100, $\alpha$=0.6). For each scenario, $c_\lambda$ is kept fixed over the replications as indicated in the second column.}
\small
\begin{center}
\begin{sc}
\vspace{0.1cm}
\begin{tabular}{rrrrr}
\Xhline{3\arrayrulewidth}
\multicolumn{1}{r|}{$\bm{n}$} & \multicolumn{1}{r|}{$\bm{c_\lambda}$} & \multicolumn{1}{@{\hskip 0.4cm}c}{\textit{glmnet}} & \multicolumn{1}{c}{\textit{sklearn}} & \multicolumn{1}{c}{\textit{ssnal-en}} \\ 
\Xhline{3\arrayrulewidth}
\multicolumn{1}{r|}{$1e4$}        & \multicolumn{1}{r|}{0.5}    
& 0.074 $(0.002)$                                 & 0.097   $(0.001)$         & \textbf{0.029}   $(0.002)$                            \\
\multicolumn{1}{r|}{$1e5$}        & \multicolumn{1}{r|}{0.6}  
       & 0.846 $(0.019)$                           & 1.170 $(0.013)$         & \textbf{0.212}  $(0.007)$              \\
\multicolumn{1}{r|}{$5e5$}        & \multicolumn{1}{r|}{0.7}   
        &  3.868 $(0.014)$                           & 5.963   $(0.462)$      & \textbf{0.789}  $(0.023)$            \\
\Xhline{3\arrayrulewidth}             
\end{tabular}
\end{sc}
\end{center}
\label{tab:standard_errors}
\vspace{-0.2cm}
\end{table}

\subsection{Different values $\mathbf{n_0}$ of $\mathbf{n_0}$, $\mathbf m$, $\mathbf{snr}$, $\mathbf \alpha$, and $\mathbf{x^*}$}

\label{app:sim_diff_values}

\begin{table}[H]
\vspace{-0.4cm}
\caption{
CPU time (in seconds) of \texttt{glmnet}, \texttt{sklearn} and \texttt{SsNAL-EN} for different simulation scenarios. Data has been generated as described in Section \ref{sec:sim_and_case}. We set $c_\la$ in order to have a solution with $n_0$ active components. The base parameters' values are: 
\vspace{0.1cm}
\newline
\centerline{
$n_0 = 5; \quad m = 500; \quad snr = 5; \quad \alpha = 0.9; \quad  x^*=5$}
\newline 
\vspace{0.1cm}
In each panel of the table, just one of these parameters is changed, as indicated in the top row. 
For \emph{SsNAL-EN}, we also report the number of iterations in parantheses.
}
\small

\begin{sc}
\centerline{
\vspace{0.1cm}
\begin{tabular}{rrrrcccccc}
\Xhline{3\arrayrulewidth}
\multicolumn{1}{l}{\textit{\textbf{}}} & \multicolumn{3}{@{\hskip 0.5cm}c}{$\bm{m=1e3}$}                                                                                        & \multicolumn{3}{@{\hskip 0.7cm}c}{$\bm{m=5e3}$}                                                                 & \multicolumn{3}{@{\hskip 0.7cm}c}{$\bm{m=1e4}$}                                                                   \\ 
\Xhline{3\arrayrulewidth}
\multicolumn{1}{r|}{$\bm{n}$}        
& \multicolumn{1}{@{\hskip 0.5cm}c}{\textit{glmnet}} & \multicolumn{1}{c}{\textit{sklearn}} & \multicolumn{1}{c}{\textit{ssnal-en}} 
& \multicolumn{1}{@{\hskip 0.5cm}c}{\textit{glmnet}}           & \textit{sklearn}           & \textit{ssnal-en}                  
& \multicolumn{1}{@{\hskip 0.5cm}c}{\textit{glmnet}}            & \textit{sklearn}            & \textit{ssnal-en}                  \\ 
\Xhline{3\arrayrulewidth}
\multicolumn{1}{r|}{$1e4$}               & 0.101                               & 0.165                                & \textbf{0.036}(5)                        & \multicolumn{1}{r}{0.727} & \multicolumn{1}{r}{1.462}  & \multicolumn{1}{r}{\textbf{0.147}(3)} & \multicolumn{1}{r}{1.481}  & \multicolumn{1}{r}{2.982}   & \multicolumn{1}{r}{\textbf{0.277}(3)} \\
\multicolumn{1}{r|}{$1e5$}               & 1.313                               & 2.099                                & \textbf{0.253}(4)                        & \multicolumn{1}{r}{7.007} & \multicolumn{1}{r}{40.159} & \multicolumn{1}{r}{\textbf{1.135}(3)} & \multicolumn{1}{r}{76.394} & \multicolumn{1}{r}{176.535} & \multicolumn{1}{r}{\textbf{1.374}(3)} \\
\multicolumn{1}{r|}{$1e6$}               & 79.395                              & 146.624                              & \textbf{2.025}(4)                        & -                         & -                          & -                                  & -                          & -                           & -                                  \\ 
\Xhline{3\arrayrulewidth}
\end{tabular}
}
\vspace{0.5cm}
\centerline{
\begin{tabular}{rrrrcccccc}
\Xhline{3\arrayrulewidth}
\multicolumn{1}{l}{\textit{\textbf{}}} & \multicolumn{3}{@{\hskip 0.5cm}c}{$\bm{snr=10}$}                                                                                        & \multicolumn{3}{@{\hskip 0.7cm}c}{$\bm{snr = 2}$}                                                                 & \multicolumn{3}{@{\hskip 0.7cm}c}{$\bm{snr = 1}$}                                                                   \\ 
\Xhline{3\arrayrulewidth}
\multicolumn{1}{r|}{$\bm{n}$}        
& \multicolumn{1}{@{\hskip 0.5cm}c}{\textit{glmnet}} & \multicolumn{1}{c}{\textit{sklearn}} & \multicolumn{1}{c}{\textit{ssnal-en}} 
& \multicolumn{1}{@{\hskip 0.5cm}c}{\textit{glmnet}}           & \textit{sklearn}           & \textit{ssnal-en}                  
& \multicolumn{1}{@{\hskip 0.5cm}c}{\textit{glmnet}}            & \textit{sklearn}            & \textit{ssnal-en}                  \\ 
\Xhline{3\arrayrulewidth}
\multicolumn{1}{r|}{$5e5$}               & \multicolumn{1}{r}{3.757}  & 5.057                                & \multicolumn{1}{r}{\textbf{0.604}(4)}                        & \multicolumn{1}{r}{4.410}  & \multicolumn{1}{r}{12.068}                               & \multicolumn{1}{r}{\textbf{0.661}(4)}                        & \multicolumn{1}{r}{4.034}  & \multicolumn{1}{r}{12.403}                               & \multicolumn{1}{r}{\textbf{0.656}(4)}                        \\
\multicolumn{1}{r|}{$2e6$}               & \multicolumn{1}{r}{78.385} & 122.110                              & \textbf{2.425}(4)                        & \multicolumn{1}{r}{75.261} & 168.902                              & \multicolumn{1}{r}{\textbf{2.437}(4)}                       & \multicolumn{1}{r}{89.667} & \multicolumn{1}{r}{131.342}                              & \multicolumn{1}{r}{\textbf{3.457}(4)}     \\   
\Xhline{3\arrayrulewidth}               
\end{tabular}
}
\vspace{0.5cm}
\centerline{
\begin{tabular}{rrrrcccccc}
\Xhline{3\arrayrulewidth}
\multicolumn{1}{l}{\textit{\textbf{}}} & \multicolumn{3}{@{\hskip 0.5cm}c}{$\bm{\alpha=0.1}$}                                                                                        & \multicolumn{3}{@{\hskip 0.7cm}c}{$\bm{\alpha=0.3}$}                                                                 & \multicolumn{3}{@{\hskip 0.7cm}c}{$\bm{\alpha=0.6}$}                                                                   \\ 
\Xhline{3\arrayrulewidth}
\multicolumn{1}{r|}{$\bm{n}$}        
& \multicolumn{1}{@{\hskip 0.5cm}c}{\textit{glmnet}} & \multicolumn{1}{c}{\textit{sklearn}} & \multicolumn{1}{c}{\textit{ssnal-en}} 
& \multicolumn{1}{@{\hskip 0.5cm}c}{\textit{glmnet}}           & \textit{sklearn}           & \textit{ssnal-en}                  
& \multicolumn{1}{@{\hskip 0.5cm}c}{\textit{glmnet}}            & \textit{sklearn}            & \textit{ssnal-en}                  \\ 
\Xhline{3\arrayrulewidth}
\multicolumn{1}{r|}{$5e5$}               & \multicolumn{1}{r}{3.341}  & \multicolumn{1}{r}{9.968}                                & \textbf{0.573}(3)                        & \multicolumn{1}{r}{3.597}  & \multicolumn{1}{r}{9.615}                                & \multicolumn{1}{r}{\textbf{0.514}(3)}                        & \multicolumn{1}{r}{3.404}  & \multicolumn{1}{r}{5.891}                                & \multicolumn{1}{r}{\textbf{0.518}(3)}                  \\
\multicolumn{1}{r|}{$2e6$}               & \multicolumn{1}{r}{90.373} & 170.966                              & \textbf{2.012}(3)                        & \multicolumn{1}{r}{89.986} & 158.160                              & \multicolumn{1}{r}{\textbf{2.030}(3)}                        & \multicolumn{1}{r}{87.147} & 129.893                              & \multicolumn{1}{r}{\textbf{2.213}(3)}      \\            
\Xhline{3\arrayrulewidth}
\end{tabular}
}

\vspace{0.5cm}
\centerline{
\begin{tabular}{rrrrcccccc}
\Xhline{3\arrayrulewidth}
\multicolumn{1}{l}{\textit{\textbf{}}} & \multicolumn{3}{@{\hskip 0.5cm}c}{$\bm{x^*=100}$}                                                                                        & \multicolumn{3}{@{\hskip 0.7cm}c}{$\bm{x^*=0.1}$}                                                                 & \multicolumn{3}{@{\hskip 0.7cm}c}{$\bm{x^*=0.01}$}                                                                   \\ 
\Xhline{3\arrayrulewidth}
\multicolumn{1}{r|}{$\bm{n}$}        
& \multicolumn{1}{@{\hskip 0.5cm}c}{\textit{glmnet}} & \multicolumn{1}{c}{\textit{sklearn}} & \multicolumn{1}{c}{\textit{ssnal-en}} 
& \multicolumn{1}{@{\hskip 0.5cm}c}{\textit{glmnet}}           & \textit{sklearn}           & \textit{ssnal-en}                  
& \multicolumn{1}{@{\hskip 0.5cm}c}{\textit{glmnet}}            & \textit{sklearn}            & \textit{ssnal-en}                  \\ 
\Xhline{3\arrayrulewidth}
\multicolumn{1}{r|}{$5e5$}               & \multicolumn{1}{r}{3.386}  & 5.529                                & \multicolumn{1}{r}{\textbf{0.545}(3)}                        & \multicolumn{1}{r}{3.403}  & \multicolumn{1}{r}{6.549}                                & \multicolumn{1}{r}{\textbf{1.154}(4)}                        & \multicolumn{1}{r}{3.423}  & \multicolumn{1}{r}{4.838}                                & \multicolumn{1}{r}{\textbf{1.405}(4)}                        \\
\multicolumn{1}{r|}{$2e6$}               & \multicolumn{1}{r}{92.001} & 151.765                              & \multicolumn{1}{r}{\textbf{2.047}(3)}                        & \multicolumn{1}{r}{66.764} & \multicolumn{1}{r}{138.826}                              & \multicolumn{1}{r}{\textbf{6.350}(5)}                        & \multicolumn{1}{r}{58.492} & \multicolumn{1}{r}{134.440}                              & \multicolumn{1}{r}{\textbf{12.109}(5)}                     \\                    
\Xhline{3\arrayrulewidth}
\end{tabular}
}
\end{sc}
\vspace{-0.2cm}
\end{table}

$\phantom{}$ \\
$\phantom{}$ \\
$\phantom{}$ \\
$\phantom{}$ \\
$\phantom{}$ \\
$\phantom{}$ \\
$\phantom{}$ \\
$\phantom{}$ \\
$\phantom{}$ \\
$\phantom{}$ 

\subsection{Screening solvers}
\label{app:sim_screening_solvers}

\begin{table}[H]
\vspace{-0.5cm}
\caption{ CPU time (in seconds) for 6 different solvers. Since \texttt{celer} and \texttt{gsr} can only solve Lasso, we tested with $\alpha=0.999$. Data has been generated as described in Section \ref{sec:sim_and_case}. In this case, for \texttt{SsNAL-EN} we start from $\sa^0 = 1$ and we increase it by a factor of 10 every iteration. Two scenarios are studied: one with $n=1$e4, $m=5$e3, $n_0=500$, and one with one with $n=5$e5, $m=500$, $n_0=100$. For each scenario, we consider 4 values of $c_\lambda$ to obtain solutions with different sparsity levels. Note, $\la^{max}$ is equal to $\lVert A^Tb\rVert_{\infty} / (m \alpha)$ for \texttt{bigasso}, $\lVert A^Tb\rVert_{\infty} / m$ for \texttt{celer}, and $\lVert A^Tb\rVert_{\infty}$ for \texttt{gsr}. } 
\small
\begin{center}
\begin{sc}

\vspace{0.1cm}
\begin{tabular}{r|r|rrrrrr}
\Xhline{3\arrayrulewidth}
\multicolumn{8}{c}{$\bm{n=1e4;} \quad \bm{m=5e3;} \quad \bm{n_0=500}$} \\ 
\Xhline{3\arrayrulewidth}
$\bm{c_\la}$   & $\bm{r}$ & \multicolumn{1}{c}{\textit{glmnet}} & \multicolumn{1}{c}{\textit{biglasso}} & \multicolumn{1}{c}{\textit{sklearn}} & \multicolumn{1}{c}{\textit{gsr}} & \multicolumn{1}{c}{\textit{celer}} & \multicolumn{1}{c}{\textit{ssnal-en}} \\ 
\Xhline{3\arrayrulewidth}
0.9 & 11  
& 0.769                      & 0.513                       & 1.449                      
& 1.336                      & 1.063                       & \textbf{0.086}                        \\
0.7 & 89  
& 1.023                       & 0.513                      & 1.484                  
& 1.227                     & 1.026                       & \textbf{0.404}                  \\
0.5 & 271  
& 0.889                      & \textbf{0.695}                        & 1.523                        
& 1.956                      & 1.201                       & 1.169                        \\
0.3 & 584 
 & 1.094                     & \textbf{1.002}                        & 1.688                        
 & 2.655                      & 1.480                       & 2.775                        \\
\Xhline{3\arrayrulewidth}                  
\end{tabular}

\vspace{0.5cm}
\begin{tabular}{r|r|rrrrrr}
\Xhline{3\arrayrulewidth}
\multicolumn{8}{c}{$\bm{n=5e5;} \quad \bm{m=500;} \quad \bm{n_0=100}$}                                                                                                                                                          \\ 
\Xhline{3\arrayrulewidth}
$\bm{c_\la}$   & $\bm{r}$ & \multicolumn{1}{c}{\textit{glmnet}} & \multicolumn{1}{c}{\textit{biglasso}} & \multicolumn{1}{c}{\textit{sklearn}} & \multicolumn{1}{c}{\textit{gsr}} & \multicolumn{1}{c}{\textit{celer}} & \multicolumn{1}{c}{\textit{ssnal-en}} \\ 
\Xhline{3\arrayrulewidth}
0.9 & 6  
& 4.607                      & 1.815                       & 4.599                       
& 7.666                      & 2.032                       & \textbf{1.351}                         \\
0.7 & 65  
& 4.537                      & 2.575                       & 6.206                    
& 10.046                      & 2.648                       & \textbf{2.005}                        \\
0.5 & 178  
& 3.964                      & \textbf{2.693}                      & 7.387                        
& 6.118                      & 3.362                      & 5.206                        \\
0.3 & 307 
 & 4.242                     & 4.736                     & 11.569                        
 & 6.392                      & \textbf{3.965}                     & 6.199                        \\
\Xhline{3\arrayrulewidth}                  
\end{tabular}
\vspace{0.5cm}

\end{sc}
\end{center}
\label{tab:screening}
\vspace{-0.2cm}
\end{table}

\subsection{Solution path}
\label{app:sim_solution_path}

\begin{table}[H]
\vspace{-0.5cm}
\caption{ CPU time (in seconds) for the computation of a solution path. Data has been generated as described in Section \ref{sec:sim_and_case} and the parameters are chosen as in \textbf{sim1} (m=500, $n_0$=100). The full $c_\la$-grid consists in 100 log-spaced points between 1 and 0.1. We truncate the path search when 100 active components are selected. $runs$ is the number of different $c_\la$ values that have been explored} 
\small
\begin{sc}
\vspace{0.2cm}
\centerline{
\scalebox{0.9}{
\begin{tabular}{r|crrrr|c|rrrr|c|rrrr}
\Xhline{3\arrayrulewidth}
\multicolumn{1}{l|}{} & \multicolumn{5}{c|}{$\bm{n=1e5}$} & \multicolumn{5}{c|}{$\bm{n=5e5}$} & \multicolumn{5}{c}{$\bm{n=1e6}$} \\ 
\Xhline{3\arrayrulewidth}
$\bm{\alpha}$ & 
\multicolumn{1}{c|}{$runs$} & \multicolumn{1}{c}{\textit{glmnet}} & \multicolumn{1}{c}{\textit{biglasso}} & \multicolumn{1}{c}{\textit{sklearn}} & \multicolumn{1}{c|}{\textit{ssnal-en}} & 
$runs$ & 
\multicolumn{1}{c}{\textit{glmnet}} & \multicolumn{1}{c}{\textit{biglasso}} & \multicolumn{1}{c}{\textit{sklearn}} & \multicolumn{1}{c|}{\textit{ssnal-en}} & 
$runs$ & 
\multicolumn{1}{c}{\textit{glmnet}} & \multicolumn{1}{c}{\textit{biglasso}} & \multicolumn{1}{c}{\textit{sklearn}} & \multicolumn{1}{c}{\textit{ssnal-en}} \\ 
\Xhline{3\arrayrulewidth}
$0.8$ & \multicolumn{1}{c|}{18} & 2.099 & 1.567 & 13.024 & $\bm{1.083}$ &
15 & 9.407 & 5.956 & 51.634 & $\bm{3.952}$ & 
16 & 22.484 & $\bm{10.732}$ & 113.641 & 13.202 \\
$0.6$ & \multicolumn{1}{c|}{17} & 1.959 & 1.583 & 9.291 & $\bm{ 0.763}$&
14 & 10.279 & 6.921 & 46.132 & $\bm{3.557}$ & 
15 & 22.548 & 11.067 & 104.541 & $\bm{6.228}$ \\ 
\Xhline{3\arrayrulewidth}
\end{tabular}
}}
\end{sc}
\label{tab:solution_path}
\vspace{-0.2cm}
\end{table}

\end{document}